%% file: main.tex
\documentclass[runningheads]{llncs}

\usepackage{eccv}

\usepackage{eccvabbrv}

\usepackage{graphicx}
\usepackage{booktabs}

\usepackage{wrapfig}
\usepackage{comment}

\usepackage{longtable}
\usepackage{placeins}

\usepackage{etoc}

\usepackage[accsupp]{axessibility}  

\usepackage{tikz,pgfplots}
\pgfplotsset{compat=1.18}
\usepackage{hyperref}
\usepackage{multirow}
\usepackage{subcaption}
\usepackage{xcolor}
\definecolor{socoBlue}{RGB}{44,105,176}
\definecolor{socoOrange}{RGB}{230,126,34}
\definecolor{socoGray}{gray}{0.45}

\usepackage{orcidlink}

\newcommand{\cmark}{\textcolor{ForestGreen}{\ding{51}}}%
\newcommand{\xmark}{\textcolor{red}{\ding{55}}}%
\newcommand{\ours}{SOCO\xspace}

\usepackage{mdframed}

\usepackage{soul}

\sethlcolor{gray!25}

\newmdenv[
  backgroundcolor=eccvblue!10,
  linecolor=gray!60!black,
  roundcorner=8pt,
  leftmargin=0pt,
  rightmargin=0pt,
  innertopmargin=3pt,
  innerbottommargin=3pt,
  innerleftmargin=3pt,
  innerrightmargin=3pt,
  skipabove=4pt,
  skipbelow=4pt
]{eccvbox}

\usepackage{pifont}
\usepackage{newunicodechar}
\newunicodechar{✓}{\checkmark}
\newunicodechar{✗}{\ding{55}}

\begin{document}

\newcommand{\papertitle}{\ours: Benchmarking Semantic Object Correspondence in Vision Foundation Models}
\title{\papertitle}

\titlerunning{\ours}

\author{
Olaf D\"unkel\inst{1}\textsuperscript{$\star$} \and
Basavaraj Sunagad\inst{2}\textsuperscript{$\star$} \and
Haoran Wang\inst{1} \and
David T. Hoffmann\inst{3} \and
Christian Theobalt\inst{1} \and
Adam Kortylewski\inst{2}
}

\authorrunning{O.~D\"unkel and B. Sunagad et al.}

\institute{Max Planck Institute for Informatics, SIC \and
CISPA Helmholtz Center for Information Security \and
University of Freiburg
}

\maketitle

\let\oldthefootnote\thefootnote
\renewcommand{\thefootnote}{\fnsymbol{footnote}}
\footnotetext[1]{Equal contribution.}
\let\thefootnote\oldthefootnote

\begin{center}
\vspace{-13pt}
{\small \textbf{Project Page:} \url{https://genintel.github.io/SOCO/}}
\vspace{6pt}
\end{center}

\input{figures/teaser_fig}
\input{sec/0_abstract}

\input{sec/1_intro2}
\input{sec/2_related_work}

\input{sec/3_methods}
\input{sec/4_exp}
\input{sec/5_conclusion}

\input{sec/X_ack}

\bibliographystyle{splncs04}
\bibliography{main}

\input{sec/X_suppl}

\end{document}

%% file: figures/teaser_fig.tex
\begin{center}
    \centering
    \captionsetup{type=figure}
    \vspace{-5pt}
    \includegraphics[width=\textwidth]{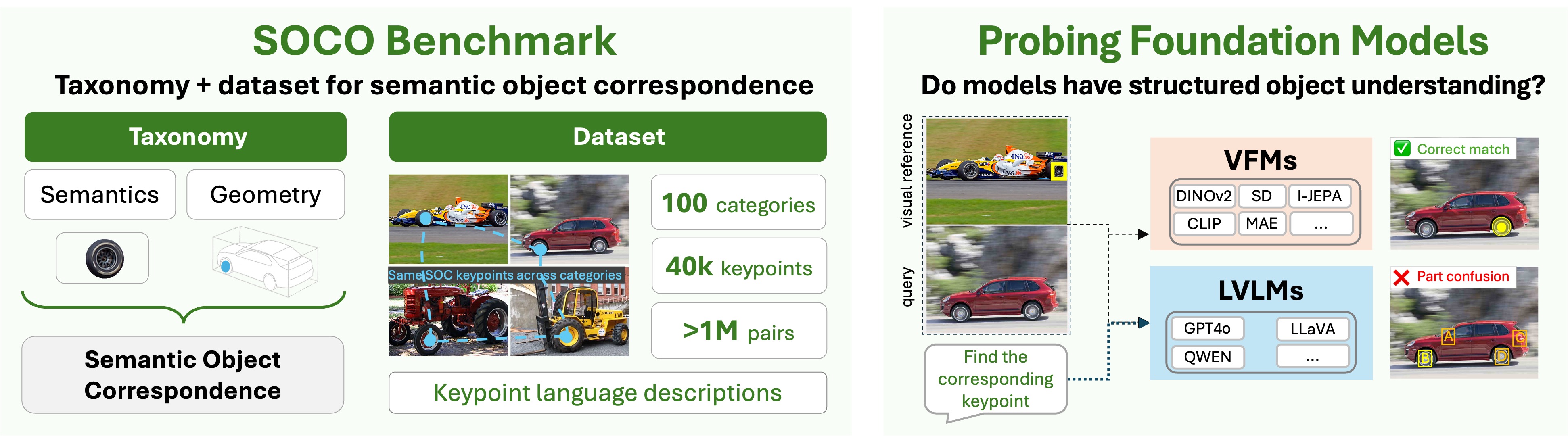}
    \vspace{-13pt}
    \captionof{figure}{
SOCO provides the first taxonomy-driven, language-grounded formulation of \textbf{Semantic Object Correspondence} (SOC), enabling structured, semantically coherent, and cross-category part annotations across 100 diverse categories,
which allows evaluating semantic and structured object understanding in vision foundation models (VFMs) and large vision language models (LVLMs).
}
    \label{fig:teaser_figure}

    \vspace{-8pt}
    
\end{center}%

%% file: sec/0_abstract.tex
\begin{abstract}
Measuring structured object understanding in vision foundation models remains challenging due to inconsistent evaluation protocols and limited part-level supervision. Semantic correspondence (SC) evaluates this capability by testing whether object parts can be matched across instances and categories under large variations in appearance, viewpoint, and geometry. To enable a systematic SC evaluation, we introduce SOCO, a new benchmark for Semantic Object Correspondence that introduces a taxonomy of correspondence types and provides consistent, functionally meaningful keypoint annotations across 100 categories and over 1M correspondence pairs. In addition, SOCO includes keypoint language descriptions, enabling the evaluation of large vision-language models (LVLMs) and their fine-grained part-level understanding. Comprehensive experiments reveal that (i) vision foundation backbones encode strong semantic structure but transfer correspondences poorly across related categories and only partially capture object-part position, (ii) LVLMs are stronger at text-prompted part localization than at visual-reference cross-image matching, exposing a gap between language-grounded localization and fine-grained visual correspondence, and (iii) correspondence performance predicts dense downstream tasks---segmentation, tracking, 3D pose estimation, and 3D detection---more strongly than ImageNet classification. Together, these findings position SOCO as a benchmark for structured, part-level representation quality in vision and multimodal foundation models.
\keywords{Semantic Correspondence \and Representation Learning \and Benchmarking}
\end{abstract}

%% file: sec/1_intro2.tex
\vspace{-.5cm}
\section{Introduction}

Visual representations form the foundation of visual intelligence. 
Evaluating their quality has long been central to progress in computer vision.
Existing benchmarks probe distinct aspects of visual understanding, 
ranging from category-level recognition benchmarks such as ImageNet~\cite{deng2009imagenet} to spatial localization tasks including detection, segmentation, and pose estimation~\cite{lin2014microsoft,cordts2016cityscapes,andriluka20142d}.
However, they provide limited insight into whether a representation captures \emph{structured object understanding}, i.e., the ability to relate semantically corresponding parts across different object instances and categories.

Recently, \textit{semantic correspondence (SC)} has become increasingly important for evaluating self-supervised and foundation models~\cite{simeoni2025dinov3,venkataramanan2025franca,ranzinger2026c,fu2024blink}, as it measures a model's ability to establish correspondences between object parts across different instances of a category---a capability that requires consistently capturing object structure under substantial variation in appearance, viewpoint, and geometry.
The ability to establish such correspondences is crucial for transferring knowledge across related objects,
for example when adapting affordances, recognition, pose estimation, or reconstruction 
to unseen categories, which is important for embodied and robotic systems \cite{florence2018dense,wang2019normalized}.

However, despite this growing adoption, progress in SC research has been constrained by the lack of a clear task definition and by the limitations of existing datasets~\cite{min2019spair,misc210k2023}.
Current benchmarks conflate \emph{two distinct abilities} in a single within-category score---recognizing the same local concept (e.g.\ a wheel center) and identifying its correct \emph{repeated instance} within an object (front-left vs.\ rear-right wheel)---and do not evaluate transfer \emph{across related categories} (a wheel center on a car, bus, or tractor) at all.
This ambiguity limits current evaluations of modern foundation models.

We therefore propose \textbf{Semantic Object Correspondence (SOC)}, a taxonomy-driven formulation of semantic correspondence that disentangles these three abilities.
SOC explicitly models the relationship between object part semantics and the overall object structure, providing a clearer separation between local concept recognition, object-relative identity, and cross-category transfer.
Concretely, the taxonomy distinguishes \emph{concept correspondence} (CC, matching the same local concept), \emph{semantic object correspondence} (SOC, matching the same concept with the same object-relative identity), and \emph{cross-category SOC} (matching object-relative keypoints across related categories through shared taxonomy concepts).
This decomposition reduces annotation ambiguity, standardizes what constitutes a valid correspondence across object categories and viewpoints, and makes distinct model failure modes separately measurable.

Building on this definition, we present \textbf{\ours}, a \textbf{S}emantic \textbf{O}bject \textbf{CO}rrespondence dataset that measures SOC with taxonomy-driven keypoint annotations across \textit{100 object categories} organized into four super-classes.
Unlike prior datasets, \ours emphasizes semantic consistency and cross-category matching, enabling structured evaluation of correspondence across varying geometry and appearance for a diverse range of man-made object and animal categories.
Across a broad family of vision foundation models---including self-supervised and vision–language models such as DINO~\cite{oquab2023dinov2,simeoni2025dinov3,caron2021emerging}, CLIP~\cite{radford2021learning}, Stable Diffusion~\cite{rombach2022high}, and I-JEPA~\cite{assran2023self}---the decomposed evaluation reveals distinct failure modes: strong VFMs recognize local concepts but exhibit large CC$\to$SOC drops (repeated-part confusion) and further SOC$\to$Cross-SOC drops (limited category-level abstraction).
Moreover, SOC is a practical zero-shot diagnostic of representation quality: it correlates with dense downstream tasks---segmentation, tracking, 3D pose, and 3D detection---more strongly than ImageNet classification accuracy.

As connecting vision and language modalities becomes increasingly important, benchmarks for structured object understanding should not only evaluate visual representations but also \textit{large vision–language models (LVLMs)}.
To support this, we extend \ours with \textit{language descriptions of correspondence keypoints}, creating a comprehensive benchmark for studying the interplay between visual correspondences and natural language in multimodal foundation models.
LVLM evaluations reveal a complementary failure mode: current LVLMs are substantially stronger at text-prompted part localization within a single image than at visual-reference correspondence across images, exposing a gap between language-grounded localization and visual matching.
Together, the results position \ours as a unified benchmark for analyzing fine-grained visual reasoning and multimodal representation quality in the era of large foundation models.

\vspace{0.5em}
\noindent\textbf{In summary, our main contributions are:}
\begin{itemize}
    \item \textbf{Task formulation.} We introduce \textbf{Semantic Object Correspondence (SOC)} as a taxonomy-driven decomposition of semantic correspondence into concept correspondence, structured object understanding, and cross-category transfer.
    \item \textbf{Dataset.} We present \textbf{\ours}, a large-scale benchmark built on this taxonomy, featuring $100$ diverse categories, semantically grounded keypoint annotations, and over $1$M correspondence pairs, with provided \textbf{language descriptions} that enable joint study of visual correspondence and language understanding in multimodal models.
    \item \textbf{Vision-model analysis.} Across a broad family of vision foundation models, the SOC decomposition exposes repeated-part confusion and limited cross-category abstraction even in strong dense self-supervised backbones.
    \item \textbf{LVLM analysis.} On the same taxonomy, current LVLMs are stronger at text-prompted part localization than at visual-reference cross-image matching, revealing a gap between language-grounded localization and fine-grained visual correspondence.
    \item \textbf{SOC as a representation diagnostic.} We conduct extensive experiments across a broad family of vision models, demonstrating that SOC correlates more strongly than ImageNet $k$NN with dense downstream tasks, positioning SOC as a practical zero-shot diagnostic of representation quality.
\end{itemize}

%% file: sec/2_related_work.tex
\section{Related work}

\textbf{Semantic Correspondence Benchmarks.}
Finding correspondences is a fundamental task in computer vision, ranging from geometric \cite{liu2008sift,rocco2017convolutional} and stereo matching \cite{scharstein2002taxonomy,mayer2016large} to optical flow \cite{butler2012naturalistic} and tracking \cite{wu2013online}, which are typically constrained to the same instance or scene.
In contrast, semantic correspondence aims to establish correspondences between object parts \textit{across} different instances of the same category.
Early datasets such as PF-PASCAL and PF-WILLOW~\cite{ham2016proposal}, TSS~\cite{taniai2016joint}, and Freiburg-Cars~\cite{freiburgcar2015iccv} defined keypoint correspondences but they were limited in scale and category diversity.
Zhang et al.~\cite{zhang2024telling} propose a semantic correspondence benchmark based on animal keypoints from AP-10K~\cite{yu2021ap}.
However, it does not include man-made objects, which have more diverse keypoint types and are equally important for probing general object-level understanding.
SPair-71k~\cite{min2019spair} became the de-facto standard benchmark by providing 71k image pairs across 1,800 images from 10 rigid categories of PASCAL 3D+~\cite{xiang2014beyond} and 8 non-rigid categories of PASCAL VOC 2012~\cite{everingham2015pascal}, out of which 481 images are used for testing.
Due to the imbalanced class selection, quadruped animals and vehicles are favored.
MISC210K~\cite{misc210k2023} focuses on multi-instance correspondence and increases dataset scale, but its keypoints are defined by geometric heuristics rather than a hierarchical taxonomy of semantic concepts, which---as in SPair-71k---prevents cross-category evaluation.
Additionally, current SC benchmarks do not provide keypoint descriptions, preventing systematic evaluation of LVLMs.
\ours addresses these limitations by introducing the concept of \textit{Semantic Object Correspondence},
a taxonomy-driven formulation that specifically separates geometric from non-geometric semantic correspondences and standardizes what constitutes a valid correspondence across object categories.
Based on this, we create a dataset of diverse categories with taxonomy-driven SC keypoints and textual descriptions, forming the basis for a more comprehensive benchmark.

\begin{table}[t]
\caption{\textbf{Comparison of semantic correspondence benchmarks.}
SOCO uniquely combines a hierarchical keypoint taxonomy, language descriptions, 
and cross-category correspondence pairs while covering a large and diverse set of categories, compared to other SC datasets that include man-made objects.
}
\vspace{-5pt}
\label{tab:related_work}
\centering
\scriptsize
\setlength{\tabcolsep}{4pt}
\begin{tabular}{lccccccc}
\toprule
Dataset & \#Cats.&\#Pairs&Keyp.  & Taxonomy  &Sep.geo.& Cross-cat. & Language \\
\midrule
PF-WILLOW~\cite{ham2016proposal}   & 5&900&10      & \xmark  &\xmark  & \xmark  &\xmark \\
PF-PASCAL~\cite{ham2016proposal}    & 20&1.3k&4--17   & \xmark  &\xmark  & \xmark  &\xmark \\
SPair-71k~\cite{min2019spair}    & 18&71k&3--30   & \xmark  &\xmark& \xmark  &\xmark \\
DISCOBOX~\cite{lan2021discobox}    & 12&36k&1--12   & \xmark  &\xmark  & \cmark  &\xmark \\
MISC210K~\cite{misc210k2023}    & 34&218k&5--52   & \xmark  &\xmark  & \xmark  &\xmark \\
\midrule
\textbf{SOCO} & \textbf{100}&\textbf{1M}&\textbf{6--32} & \textbf{\cmark} & \textbf{\cmark} & \textbf{\cmark}  &\textbf{\cmark}  \\
\bottomrule
\end{tabular}
\vspace{-10pt}
\end{table}

\textbf{Semantic Correspondence in the Era of Foundation Models.}
Self-supervised and multimodal foundation models have renewed interest in semantic correspondence as a probe for representation quality~\cite{el2024probing,venkataramanan2025franca,simeoni2025dinov3}, after various studies have shown that features obtained from such models can be utilized for identifying semantic correspondences in a zero-shot manner~\cite{caron2021emerging, tang2023emergent,oquab2023dinov2,zhang2023tale,stracke2025cleandift,gan2026unleashing,luo2023diffusion}, even though they do not encode the 3D part composition particularly well~\cite{mariotti2024improving,zhang2024telling,sommer2025common3d,chic3po,dunkel2025yourself,mariotti2025jamais,wandel2025semalign3d}.
Evaluating semantic correspondence (SC) performance provides a complementary diagnostic to conventional tasks such as classification~\cite{deng2009imagenet,dunkel2025cns,everingham2010pascal} or segmentation~\cite{zhou2017scene,cordts2016cityscapes}: 
by measuring how well models align object parts under appearance and pose variation, it reveals whether representations encode \textit{fine-grained part-level} and \textit{3D-aware} structure rather than local appearance details or global category cues.

In parallel to advances in SSL, vision–language models (VLMs) such as CLIP~\cite{radford2021learning}, BLIP~\cite{li2022blip}, and Flamingo~\cite{flamingo2022} were developed to align visual and textual modalities, but their evaluation focuses mainly on retrieval and captioning~\cite{radford2021learning, shen2022how} rather than fine-grained spatial understanding.
Moreover, modern large vision-language models (LVLMs) such as LLava~\cite{liu2023llava}, Qwen-VL~\cite{Qwen-VL}, GPT-4V~\cite{openai2023gpt4v}, and Gemini~\cite{geminiteam2025geminifamily}, extend this paradigm toward multimodal visual reasoning, yet their evaluation remains dominated by high-level tasks like VQA~\cite{yue2023mmmu, liu2024mmbench} and high-level spatial reasoning~\cite{yang2024vsibench}.
BLINK~\cite{fu2024blink} contains a limited number of questions targeting semantic correspondence. 
However, since it is built on SPair-71k and does not contain language annotations, this benchmark does not provide a comprehensive evaluation of diverse fine-grained object understanding.
Our work addresses this gap by introducing a benchmark that enables a systematic evaluation of LVLMs in terms of their visual correspondence and natural language alignment, allowing analysis of how linguistic cues influence fine-grained correspondence-level understanding.

%% file: sec/3_methods.tex
\section{A Taxonomy for Semantic Correspondence}

Semantic correspondence (SC) is commonly understood as the task of matching points with similar semantics across different instances of an object category. 
However, the definition of ``semantic'' correspondence has remained vague and dataset-dependent.
We detail this in \cref{sec:corr:limitations}.
To address this gap, we propose a taxonomy for the SC task, 
providing a principled foundation for systematic annotation and evaluation.
The proposed taxonomy forms the conceptual basis for \textbf{Semantic Object Correspondence (SOC)}, 
a formulation that explicitly separates the local semantics and geometric position of an object part.
We introduce SOC in the following section (\cref{sec:corr:taxonomy}) and show how it resolves the inconsistencies observed in existing benchmarks.

\subsection{Limitations of Current SC Keypoint Annotations}\label{sec:corr:limitations}

Existing SC object datasets (e.g., PF-PASCAL~\cite{ham2016proposal}, MISC210K~\cite{misc210k2023}, Freiburg Cars~\cite{freiburgcar2015iccv}, SPair-71k~\cite{min2019spair}) lack a systematic, hierarchical keypoint annotation strategy that scales across categories.
Their annotations are often defined geometrically (e.g., midpoints on TV or boat contours) rather than as self-contained semantic concepts, are ambiguous for categories with large intra-class variability (\textit{boats}) or symmetry (\textit{bottles}, \textit{potted plants}), are defined on 2D projections and thus break under viewpoint change, and are sometimes internally inconsistent (e.g., the ``end'' of a \textit{train}).
Crucially, current benchmarks evaluate object correspondence only \emph{within} categories, ignoring relationships between semantically related objects (cars/trucks/buses) and thereby preventing assessment of cross-category semantic transfer.
We illustrate concrete cases of annotation limitations in \cref{fig:spair_problems} in the supplementary.

These limitations are not merely annotation artifacts but stem from the absence of a structured representation of object parts.
A principled formulation requires three properties: keypoints grounded in local semantics (unambiguous identification); \emph{identity attributes} that distinguish repeated parts (front-left vs.\ rear-right wheel); and an explicit hierarchical organization of semantic concepts that can be reused across categories rather than redefined per class.

\subsection{A Taxonomy of Semantic Object Correspondence}\label{sec:corr:taxonomy}
To introduce a more principled formulation of semantic correspondence, we define the term \textbf{Semantic Object Correspondence (SOC)}. 
SOC explicitly separates two complementary aspects: the local semantics of an object part and its spatial configuration within the overall object structure.
This allows probing whether a model is able to match \textit{semantic concepts} and \textit{semantic object keypoints} that include a positional attribute.

\begin{figure}[t]
    \centering
    \includegraphics[width=\linewidth]{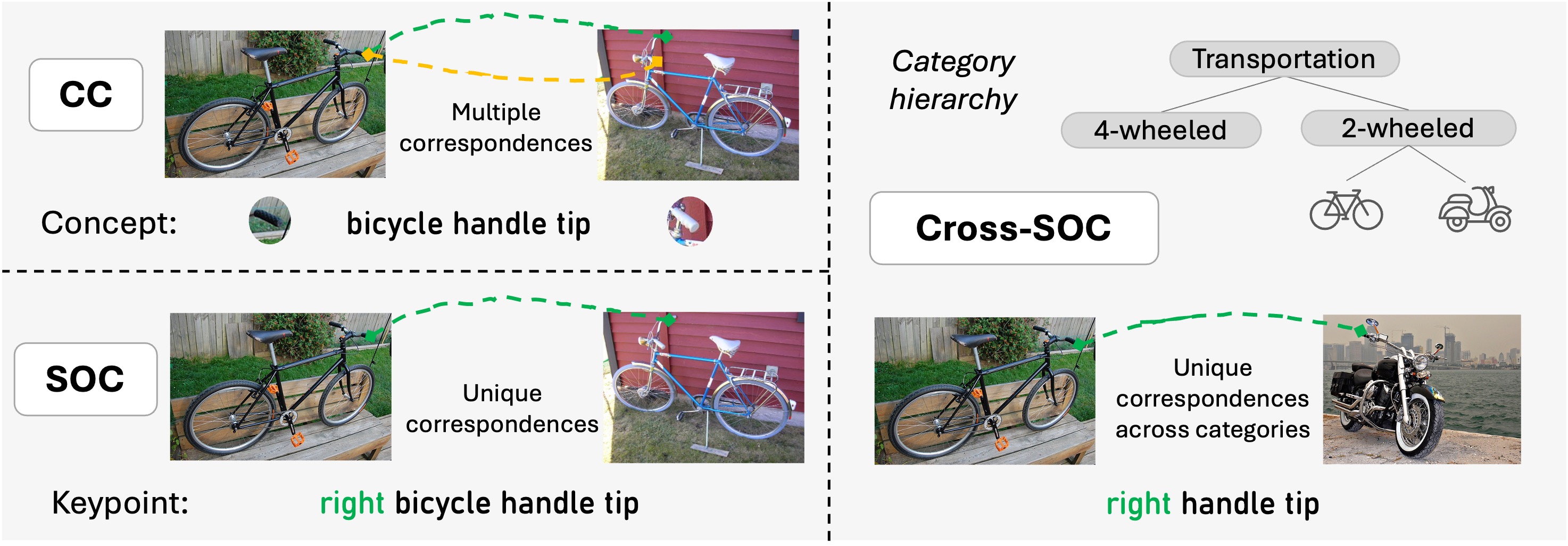}
    \vspace{-10pt}
    \caption{\textbf{Illustration of concept correspondence (CC), semantic object correspondence (SOC), and cross-category semantic object correspondence (Cross-SOC).} 
    SOCO differentiates CC and SOC, which define unique correspondences by disambiguating multiple instances of the same concept via geometric attributes, such as \textit{right}.
    Cross-category matches (Cross-SOC) are derived from the accompanying category hierarchy.
    }
    \label{fig:illustration_cc_soc}
    \vspace{-15pt}
\end{figure}

A \textbf{semantic concept} is defined as a uniquely identifiable location (e.g., a corner point) within an object part that is typically shared across instances of the same category.
Concepts capture the local semantics of a location on an object and its immediate functional context—for instance, the \textit{door handle} of a car, irrespective of whether it belongs to the left or right door.
In contrast, \textbf{semantic object keypoints} are concrete, instance-specific realizations of a semantic concept.
Each semantic object keypoint inherits from a concept but is further disambiguated by additional positional attributes that describe its placement within the object or component, such as \textit{left}, \textit{right}, \textit{bottom}, or \textit{rear}, which are consistently defined in the object-centric coordinate system.
Concepts therefore describe \emph{what} object part is being matched, whereas semantic object keypoints also specify \emph{which instance} of that part within an object is considered.
This makes finding correspondences among semantic object keypoints more challenging than concept-level matching,
as a model must capture or reason about both semantic identity and geometric placement within the object context to correctly identify correspondences.
While matching \textit{concepts} across two instances (\textit{concept correspondence, CC}) can yield non-unique matches across keypoints, keypoint matching (\textit{semantic object correspondence, SOC}) always has a unique solution.
Formally, a \textit{Semantic Object Correspondence} is defined as a match between two keypoints that share the same semantic concept and identical object-relative identity attributes, ensuring both semantic and geometric matching.

Importantly, semantic concepts are not restricted to a single category
(\textit{cross-category SOC or Cross-SOC}).
For example, a \textit{wheel} concept may appear in a passenger car and in a school bus or tractor.
To capture this hierarchical and cross-category structure, we propose to organize all semantic concepts within a taxonomy that spans categories, super-categories, and shared concepts among objects.
This hierarchy enables correspondence evaluation both within categories and across related object classes.
\cref{fig:illustration_cc_soc} illustrates CC, SOC, and Cross-SOC.

Together, this formulation establishes a coherent and extensible annotation framework for semantic correspondence, which forms the conceptual foundation for the \textbf{\ours} dataset introduced next.

\section{The SOCO dataset}

Building on the taxonomy introduced in \cref{sec:corr:taxonomy}, 
we construct \textbf{SOCO}: a large-scale, taxonomy-driven dataset
for evaluating \textit{Semantic Object Correspondence (SOC)}.
SOCO is designed to address key limitations of prior correspondence benchmarks by providing 
(1) a standardized, semantically grounded keypoint schema, 
(2) cross-category and hierarchical image-keypoint pairs,
(3) and a substantially broader and more balanced set of object categories.
Additionally, \ours introduces \textit{language descriptions} for all keypoints, enabling unified evaluation of both vision and vision–language correspondence models.

\begin{figure*}[t]
    \centering
    \includegraphics[width=\linewidth]{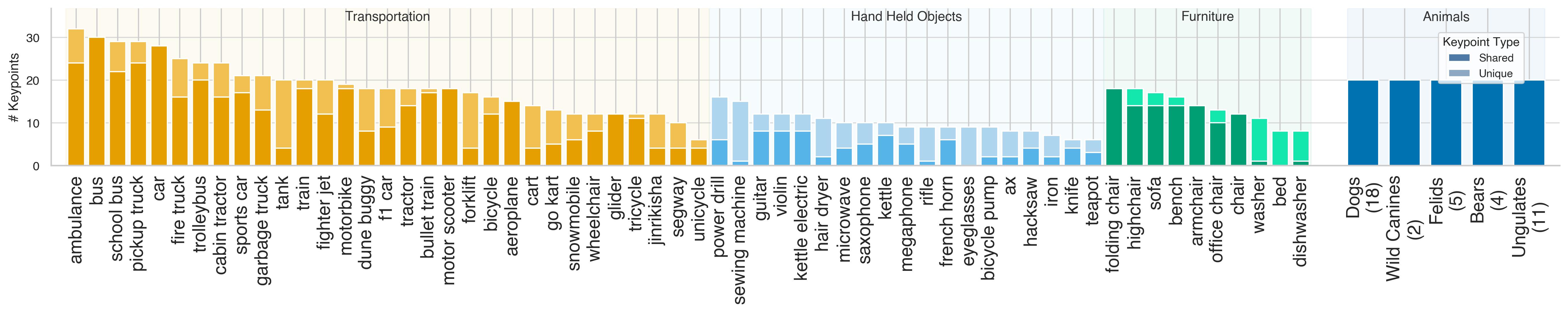}
    \vspace{-15pt}
    \caption{\textbf{Statistics of labeled keypoints.}
    Keypoints in \ours are annotated for a diverse set of categories from four super-categories.
    Each category is labeled with a subset of keypoints that are shared across multiple categories.
    The animal keypoints are shared across all animal categories.
    }
    \label{fig:dataset_statistics}
    \vspace{-15pt}
\end{figure*}

\subsection{Dataset Creation}
In the following, we describe the steps of the dataset creation: 
Image collection, category distribution, keypoint annotation, and language descriptions.

\textbf{Image collection.}
All images are samples from ImageNet.
We rely on 2D and 3D annotations from ImageNet3D~\cite{ma2024imagenet3d} for man-made objects and on keypoint annotations from the Animal3D dataset~\cite{xu2023animal3d} for the animal categories.
We only retain images that (1) contain valid pose metadata, (2) depict a single salient object, and (3) have a sufficiently large object size.

\textbf{Category distribution.}
SOCO comprises \textbf{100 categories} organized into four high-level super-categories:
\textit{Transportation} (31 classes), \textit{Hand-held Objects} (20 classes), \textit{Furniture} (9 classes), and \textit{Animals} (40 classes).

\textbf{Keypoint annotation.}
All keypoints follow the introduced taxonomy.
While annotations for animal categories can be acquired from animal keypoint datasets, annotations of man-made objects that follow the taxonomy do not exist and, therefore, need to be collected.
For this purpose, initial annotations are acquired via Amazon Mechanical Turk and refined through a manual verification stage.
A user-friendly UI with integrated keypoint reference cards was developed to enable high-quality annotations. 
Three qualified annotators independently complete each image annotation, and the annotations are median-aggregated after removing outliers.
Every keypoint annotation is verified manually to ensure consistency and accuracy.
The median per-keypoint standard deviation across annotators is 0.85\% (normalized by the maximum image dimension), indicating strong agreement.
During manual verification, 65.4\% of annotations required only minor refinements within PCK@0.05 tolerance, while 6.8\% required larger corrections, e.g., due to confused conventions (e.g. left vs.\ right).

\textbf{Language Descriptions.}
Each annotated keypoint includes a human-specified language description that combines its categorical, conceptual, and geometric attributes. 
Descriptions are generated programmatically using the tuple 
\small\texttt{(category, concept, keypoint position within the object part, object part position within the whole object)},
\normalsize
e.g., \textit{“Center point of the front left wheel of a bus”}.

\subsection{Dataset Statistics}

\Cref{fig:dataset_statistics} presents the per-category keypoint distribution, including how many keypoints are shared with other categories.
For each object category, 40 images are annotated, ensuring diverse viewpoints, shapes, and instance-level variations, resulting in a total of 4000 images.
We construct Semantic Object Correspondence (SOC) pairs by matching images within the same category, requiring at least three shared semantic keypoints. 
This yields around 73k SOC pairs with a total of around 560k keypoint correspondences. 
Concept correspondences (CC) are generated using the same pairs.

We also form cross-category (Cross-SOC) pairs, using a minimum of three shared semantic keypoints. 
Due to the large combinatorial space of cross-category pairings, Cross-SOC generation results in around 1.3M cross-category correspondence pairs. 
These complementary pairing regimes (CC, SOC, and Cross-SOC) provide progressively more challenging correspondences that support evaluation across concepts, keypoints, and different categories.

%% file: sec/4_exp.tex
\section{Experiments}

In this section, we benchmark several foundation models on semantic object correspondence.
We first report results for vision encoders (\cref{sec:soco_eval}) and LVLMs (\cref{vlm_results}).
Then, we analyze how SOC relates to other vision tasks (\cref{sec:exp:soc_down}).

\subsection{Vision Foundation Model Evaluation on \ours}\label{sec:soco_eval}

\input{tables/main_results}

\textbf{Evaluation Setup.}
In the following, we evaluate common foundation models on \ours and compare their performance on three subtasks: 
\textbf{First}, \textit{concept correspondence} (CC) evaluates whether semantic concepts can be localized correctly.
\textbf{Second}, \textit{semantic object correspondence} (SOC) evaluates whether a model also encodes the geometric position of such a semantic concept relative to the whole object.
\textbf{Third}, the most challenging cross-category setting Cross-SOC probes whether representations robustly encode the evaluated concepts across different object categories.
We evaluate on three fixed random SOCO subsets, which are released together with the full dataset. 
For each task (CC, SOC, Cross-SOC), we use 20k pairs with a uniform number of image pairs per category, ensuring high category and image diversity while keeping a manageable evaluation cost.

We select a representative set of current representation learning approaches:
Self-supervised models like the DINO family~\cite{caron2021emerging,oquab2023dinov2,simeoni2025dinov3}, iBOT~\cite{zhou2021ibot}, I-JEPA~\cite{assran2023self}, MAE~\cite{he2022masked}, and PIXIO~\cite{yang2025pursuit}, vision models trained with text supervision~\cite{radford2021learning,bolya2025perception,Qwen2.5-VL} and with a multi-view reconstruction objective~\cite{croco_v2}, a generative image diffusion model~\cite{tang2023emergent,rombach2022high}, and distilled models~\cite{Ranzinger_2024_CVPR,heinrich2025radiov25improvedbaselinesagglomerative,sariyildiz2025dune}.

Following common practice in previous work \cite{zhang2023tale,simeoni2025dinov3,el2024probing}, we evaluate SOC in a zero-shot manner:
Given a source image $I^s$, a target image $I^t$, and a query point $p_i^s \in \mathbb{R}^2$ in the source image, the corresponding target point $p_i^t \in \mathbb{R}^2$ is computed by selecting the nearest feature vector in the target image through the argmax cosine similarity between the feature vector $f_i^s$ at the query point and the feature map $\mathcal{F}^\text{t}$ of the target image:
\begin{equation} \label{eq:nn_feat}
    p_i^t =\arg\max_{q_i^t \in I^t} \text{sim}\!\left( f_i^s,\mathcal{F}^t(q_i^t) \right).
\end{equation}

For model evaluation, we follow common practice~\cite{min2019spair,zhang2024telling} and evaluate the matching performance via the Percentage of Correct Keypoints (PCK).
It is defined by the ratio of correctly predicted keypoints that are within a radius of $R=\alpha \cdot\max (h,w)$ around the correct ground truth keypoint, where $h$ and $w$ refer to the height and width of the bounding box of the considered object, respectively.
In the main paper, we report PCK at $\alpha=0.1$, averaged over all image pairs of the dataset (\textit{per-img}). 
We keep the image resolution fixed across models for fair comparison with small deviations if the specified image resolution is not a multiple of the given patch size.
This benefits models with a smaller patch size. However, using a fixed number of patches does not substantially change model rankings, as we show in the supplementary.

As SOCO supports the explicit separation of geometric attributes and semantic concept, we further evaluate \textit{SOC-geo}:
This evaluates specifically whether a models is capable of differentiating the geometric positions of keypoints of the same concept.
Given one source keypoint, the argmax is computed over all instances of the same concept for a target image of the same category.
We only select image pairs where there are at least two pairs to match, which results in around $100k$ evaluated keypoint pairs.
The random performance is 41.24\% for this setting: The number of evaluated keypoints varies across categories and images. E.g., a car wheel might appear two or three times on an image but for a chair all four legs are typically visible.

\textbf{Experimental Results.}
Model evaluation results are presented in \cref{tab:soc_performance} and we summarize the findings below.

\begin{eccvbox}
Strong semantic representations in vision foundation models do not imply geometric part awareness.
\end{eccvbox}
This finding is indicated by the consistent and substantial performance drops from CC to SOC for all evaluated models.
Notably, the magnitude of this drop scales with overall model performance, suggesting that stronger semantic representations do not close the gap to geometric part awareness.
This effect persists even for the best models (e.g., DINOv2): they capture semantic concepts well but struggle to disambiguate repeated object parts, as their representations do not reliably encode object-level geometry.
Performance drops further in the cross-category setting Cross-SOC, as the appearance across labeled object parts changes even more strongly.

The per-supercategory results in \cref{tab:soc_performance} indicate substantially different performance across categories.
Further, the CC$\to$SOC gap varies: 
it is largest for \textit{Furniture} (DINOv2: SOC $45.5$ vs CC $77.5$) and \textit{Transportation}, where repeated symmetric parts, such as chair/table legs and the front/rear, left/right wheels of vehicles—dominate, and smaller for the more articulated but less repetitive \textit{Animals} and the heterogeneous \textit{Hand-held} super-categories.
Interestingly, model rankings change with object structure: 
DINOv3 outperforms DINOv2 on \textit{Furniture} ($59.9$ vs $45.5$) despite being weaker on average, and SD\,2.1 and DUNE also become comparatively stronger when repeated parts dominate.

The \textit{SOC-geo} evaluations in \cref{tab:soc_performance} , explicitly separating geometric and semantic object part localization, indicate that ranking changes between different models.
Similar to the previous observations for the furniture category, DINOv3 outperforms DINOv2 and SD2.1 achieves the best performance, which indicates that object part position is more effectively encoded here.

A single average score therefore hides \textit{which} capability a model is missing—exactly the diagnostic value our taxonomy is designed to expose.

\begin{eccvbox}
Dense self-supervised learning objectives lead to stronger semantic correspondence representations than global alignment objectives.
\end{eccvbox}
Representations from the DINO model family perform particularly well for concept correspondence (CC), indicating that their self-supervised objectives learn robust local semantic features.
DINOv2 shows clear gains over DINOv1, whereas DINOv3 performs slightly worse across all correspondence settings.
Models such as C-RADIOv3 and DUNE, which are distilled from strong dense feature encoders including DINOv2, inherit these properties and achieve competitive performance.
In contrast, models trained with global alignment objectives, such as CLIP~\cite{radford2021learning}, perform substantially worse, reflecting the limited spatial precision of their representations.
Compared with CLIP, the larger-scale PerceptionEncoder~\cite{bolya2025perception} improves correspondence performance in its spatial variant, consistent with the \ours evaluation results.
Interestingly, the vision encoder of Qwen2.5-VL~\cite{Qwen2.5-VL} performs similarly poorly to CLIP on this task.

Finally, reconstruction-based models such as MAE and CroCoV2 perform poorly, as their objectives primarily encourage instance-specific appearance reconstruction rather than semantic feature alignment.
However, PIXIO demonstrates that scaling reconstruction-based objectives can substantially improve dense correspondence representations.
I-JEPA achieves comparatively strong performance despite being trained only on ImageNet-1k.

\subsection{LVLM evaluation on SOC}\label{vlm_results}
\input{tables/vlm_results}
In this section, we analyze several representative LVLMs on SOC and compare their performance in settings with and without access to textual descriptions.

\textbf{Experimental Setup.} 
Following the BLINK benchmark~\cite{fu2024blink}, we formulate semantic correspondence as a multiple-choice VQA task. 
We adopt the \emph{CircularEval} protocol~\cite{liu2024mmbench}, where each question is presented to the LVLM four times with different permutations of the answer choices (ABCD) to enforce a consistent prediction. 
An answer is considered correct only if the model predicts the correct option in all permutations, and we report accuracy under this strict criterion.

Unlike BLINK, which uses only an annotated reference image as the visual prompt, we further study how different query prompt types affect semantic correspondence performance.
Specifically, we evaluate three settings: \textit{Vis.}, \textit{Vis.+Desc.}, and \textit{Desc.}
In all three settings, the target image with candidate keypoint markers A/B/C/D is shown to the LVLM; the settings differ only in how the query keypoint is specified (\cf the inset of Tab.~\ref{table:vlm_res}).
In \textit{Vis.}, the query is provided as a visual prompt, where the query keypoint is marked with visual markers in the source image.
In \textit{Vis.+Desc.}, a textual description of the query keypoint is provided additionally.
In \textit{Desc.}, the source image is omitted, and the query keypoint is specified using only the textual description.
The target image and its A/B/C/D candidate markers remain visible.
This results in a \textit{Random++} baseline performance of 25\%, where the predictions are consistently permuted.
The gap between \textit{Random++} and \textit{Vis.} therefore quantifies cross-image visual matching, while \textit{Desc.} measures text-prompted keypoint localization in the target image.
Full prompts and additional illustrations are provided in the supplementary.
We evaluate the LVLMs on a smaller subset of SOCO with 20 image pairs per category, and adapt DINOv2 to the same 4-choice protocol by selecting the candidate patch with the highest cosine similarity to the query feature.
The quantitative results on \ours are summarized in Tab.~\ref{table:vlm_res}.
As the evaluation follows a circular protocol, the \textit{Random++} baseline returns a random answer that is consistent across the four permuted questions of the same evaluation.

\textbf{Experimental Results.} 
LVLM evaluation results are presented in \cref{table:vlm_res}, and findings are discussed below.
\begin{eccvbox}
LVLMs are stronger at text-prompted keypoint localization than at visual-reference cross-image matching, exposing a gap between language-grounded localization and fine-grained visual correspondence.
\end{eccvbox}
A consistent trend across LVLMs is that providing an explicit keypoint description (\textit{Vis.+Desc.} and \textit{Desc.}) improves performance compared to a purely visual query (\textit{Vis.}).
Notably, all models achieve higher accuracy in the description-only setting (\textit{Desc.}) than in the visual-reference setting (\textit{Vis.}).
This indicates that LVLMs are more effective at localizing a textually described part within a single image than at transferring a marker from a source image to the target image.

Overall, recent models show clear improvements in both visual and language understanding.
For example, the Qwen family shows consistent gains from smaller to larger models, and the Qwen3-VL-8B model outperforms its Qwen2.5-VL-7B predecessor, indicating that scaling and improved training pipelines translate into stronger semantic correspondence capabilities.

However, DINOv2 adapted to the 4-choice setting outperforms all evaluated LVLMs, achieving 81.0\% compared to the best evaluated LVLM at 54.0\%.
This suggests that current LVLMs rely heavily on textual guidance but remain limited in their ability to align visual and textual modalities for fine-grained, cross-image correspondences.
Therefore, despite recent progress, semantic correspondence on \ours remains a challenging task for current LVLMs.

\subsection{Relation to Other Vision Downstream Tasks}\label{sec:exp:soc_down}

The previous sections evaluated SOC across a diverse set of models.
In contrast, vision foundation models are typically assessed on various downstream tasks~\cite{venkataramanan2025franca,oquab2023dinov2,simeoni2025dinov3,Ranzinger_2024_CVPR,bolya2025perception}, spanning global objectives (e.g., image classification) and dense prediction tasks (e.g., tracking and semantic segmentation).
These tasks require different evaluation protocols, such as linear probing for ImageNet~\cite{oquab2023dinov2}, task-specific fine-tuning, or DPT-based training~\cite{ranftl2021vision,simeoni2025dinov3}, and their outcomes can depend strongly on hyperparameter choices.

Currently, ImageNet still remains the gold standard task for measuring representation quality \cite{oquab2023dinov2, simeoni2025dinov3, assran2023self}, as it correlates well to other tasks \cite{kornblith2019better}.
However, Bolya et al. \cite{bolya2025perception} have shown that capturing global representations is not necessarily aligned with strong dense semantic features.

As SOC probes dense semantic and geometric features, it is more indicative of structured visual understanding than classification-based metrics such as ImageNet kNN, while remaining practical through a simple zero-shot protocol without hyperparameter tuning.
We therefore study its relation to other semantic and geometric vision tasks to assess whether it can serve as a representative diagnostic of representation quality.

\textbf{Experimental Setup.}
We evaluate the representational quality of modern vision and vision–language backbones on a representative set of tasks using a unified experimental protocol that builds directly on Probe3D~\cite{el2024probing}. 
We extend Probe3D with additional probes, including semantic object correspondence on SOCO, semantic segmentation \cite{zhou2017scene}, tracking \cite{doersch2022tap}, 3D pose estimation \cite{ma2024imagenet3d}, and 3D object detection using an adapted version of the Omni3D~\cite{brazil2023omni3d} pipeline.
Furthermore, we integrate a diverse set of vision foundation models, enabling performance evaluation at large scale. 
This unified design allows us to evaluate both fine-grained and object-level 3D understanding under identical backbone, decoding, and optimization conditions.
We evaluate \textit{depth estimation} and \textit{surface normal prediction} on NYU \cite{nyudepthECCV12}, \textit{geometric multi-view correspondence} on NAVI \cite{jampani2023navi}, k-nearest neighbor \textit{kNN classification} on ImageNet \cite{deng2009imagenet}, \textit{3D pose regression} on ImageNet3D \cite{ma2024imagenet3d}, \textit{semantic segmentation} on ADE-20k \cite{zhou2017scene}, and \textit{zero-shot tracking} on TAP-Vid~\cite{doersch2022tap}, covering a wide spectrum of monocular single- and multi-view spatial reasoning requiring semantic and/or geometric understanding. 
We largely follow the hyperparameters used by El Banani et al. \cite{el2024probing} and discuss implementation details in the supplementary.
We open-source the evaluation framework.

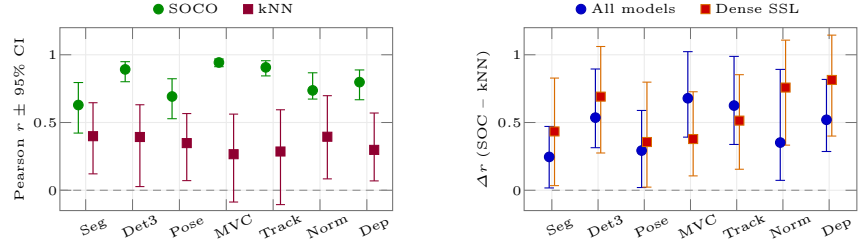
\begin{figure}[t]
  \centering
  \begin{subfigure}[c]{0.49\linewidth}
    \centering
    \input{figures/cor_with_r}
  \end{subfigure}\hfill
  \begin{subfigure}[c]{0.49\linewidth}
    \centering
    \input{figures/delta_cor}
  \end{subfigure}
  \vspace{-3pt}
  \caption{
  \textbf{Per-task Pearson $r$ across 37 vision models, with 95\% bootstrap CIs.}
  \textit{Left:} SOC correlates with every downstream task more strongly than ImageNet kNN.
  \textit{Right:} the SOC advantage $\Delta r = r_{\text{SOC}} - r_{\text{kNN}}$ stays positive on all tasks and is preserved on a 17 subset only including models trained with dense SSL objectives.
  }
  \label{fig:knn_vs_soco_corr}
  \vspace{-16pt}
\end{figure}

\textbf{Experimental Results.}
We compute the Pearson correlation between SOC performance and the downstream metrics across $37$ vision models, with 95\% bootstrap CIs (10k resamples) and leave-one-out checks.
The results are summarized in \cref{fig:knn_vs_soco_corr}.
\begin{eccvbox}
SOC has a stronger correlation to various dense geometric and semantic tasks than kNN ImageNet classification.
\end{eccvbox}
SOC dominates kNN on every evaluated downstream task (\cref{fig:knn_vs_soco_corr}), with CIs that exclude zero for the six conclusive metrics.
The advantage of SOC over kNN persists after restricting the pool to $17$ dense-SSL models, ruling out a dense-vs-global confound.
Leave-one-out resampling agrees with the full-pool results on every metric.
Overall, this suggests that SOC is a practical zero-shot diagnostic that is more aligned with dense vision tasks than ImageNet kNN.

%% file: tables/main_results.tex
\begin{table}[tbp]
\centering
\caption{\textbf{Model performances on SOCO.}
We report PCK@0.1 across concept correspondence (CC), semantic object correspondence (SOC), its cross-category variant (Cross-SOC), and the geometric variant (SOC-geo) as well as supercategory-wise results.
As more geometric awareness is required for SOC and more semantic abstraction for Cross-SOC, model performance drops for all models.
Model rankings change for geometric SOC, indicating that models encode object geometry and semantics to different extents.
Additional evaluations are provided in the supplementary.
}
\label{tab:soc_performance}

\scriptsize
\renewcommand{\arraystretch}{1.05}
\setlength{\tabcolsep}{4.0pt}

\begin{tabular}{lcccccccc}
\toprule
\textbf{Model}
&
\multicolumn{4}{c}{\textbf{Accuracies by Tasks}}
&
\multicolumn{4}{c}{\textbf{SOC by Supercategories}}
\\
\cmidrule(lr){2-5}
\cmidrule(lr){6-9}
&
\textbf{CC}
&
\textbf{SOC}
&
\textbf{Cross-SOC}
&
\textbf{SOC-geo}
&
\textbf{Trans.}
&
\textbf{Hand}
&
\textbf{Furn.}
&
\textbf{Animals}
\\
\midrule

DINOv1
& 43.8
& 30.6{\tiny\textcolor{gray}{$\downarrow$13.2}}
& 23.9{\tiny\textcolor{gray}{$\downarrow$19.8}}
& 68.4
& 29.1 & 32.9 & 27.4 & 31.4
\\

DINOv2
& \textbf{78.9}
& \textbf{60.4}{\tiny\textcolor{gray}{$\downarrow$18.5}}
& \textbf{54.9}{\tiny\textcolor{gray}{$\downarrow$24.0}}
& 71.9
& \textbf{56.9} & \textbf{61.6} & 45.5 & \textbf{66.3}
\\

DINOv3
& 69.7
& 55.5{\tiny\textcolor{gray}{$\downarrow$14.2}}
& 49.4{\tiny\textcolor{gray}{$\downarrow$20.3}}
& 76.0
& 51.6 & 57.4 & \textbf{59.9} & 56.6
\\

iBOT
& 55.2
& 39.6{\tiny\textcolor{gray}{$\downarrow$15.5}}
& 34.1{\tiny\textcolor{gray}{$\downarrow$21.1}}
& 69.4
& 36.1 & 40.1 & 32.8 & 43.9
\\

I-JEPA
& 60.5
& 46.3{\tiny\textcolor{gray}{$\downarrow$14.2}}
& 38.4{\tiny\textcolor{gray}{$\downarrow$22.1}}
& 71.7
& 41.5 & 51.0 & 46.7 & 47.7
\\

C-RADIOv3
& 69.0
& 51.1{\tiny\textcolor{gray}{$\downarrow$18.0}}
& 46.3{\tiny\textcolor{gray}{$\downarrow$22.7}}
& 68.9
& 51.7 & 48.1 & 39.7 & 54.8
\\

DUNE
& 60.1
& 45.7{\tiny\textcolor{gray}{$\downarrow$14.4}}
& 39.4{\tiny\textcolor{gray}{$\downarrow$20.7}}
& 73.5
& 40.0 & 50.5 & 51.0 & 46.6
\\

SD 2.1
& 60.5
& 49.0{\tiny\textcolor{gray}{$\downarrow$11.5}}
& 42.7{\tiny\textcolor{gray}{$\downarrow$17.8}}
& \textbf{76.8}
& 46.4 & 47.0 & 50.9 & 50.9
\\

CroCov2
& 15.1
& 10.2{\tiny\textcolor{gray}{$\downarrow$5.0}}
& 7.8{\tiny\textcolor{gray}{$\downarrow$7.3}}
& 67.2
& 11.6 & 12.3 & 10.2 & 8.1
\\

MAE
& 14.4
& 9.4{\tiny\textcolor{gray}{$\downarrow$5.0}}
& 7.1{\tiny\textcolor{gray}{$\downarrow$7.3}}
& 66.5
& 10.0 & 11.9 & 11.7 & 7.1
\\

PIXIO
& 49.5
& 37.5{\tiny\textcolor{gray}{$\downarrow$11.9}}
& 32.9{\tiny\textcolor{gray}{$\downarrow$16.5}}
& 71.8
& 37.3 & 40.2 & 46.7 & 34.2
\\

CLIP
& 25.2
& 16.2{\tiny\textcolor{gray}{$\downarrow$9.0}}
& 11.4{\tiny\textcolor{gray}{$\downarrow$13.8}}
& 64.7
& 17.9 & 14.7 & 11.7 & 16.8
\\

PE-Spatial
& 63.2
& 45.9{\tiny\textcolor{gray}{$\downarrow$17.4}}
& 40.9{\tiny\textcolor{gray}{$\downarrow$22.4}}
& 68.5
& 46.5 & 43.4 & 38.7 & 48.4
\\

QWEN-L
& 36.8
& 26.0{\tiny\textcolor{gray}{$\downarrow$10.7}}
& 22.8{\tiny\textcolor{gray}{$\downarrow$14.0}}
& 66.1
& 28.1 & 28.4 & 29.3 & 22.5
\\

\bottomrule
\end{tabular}
\vspace{-10pt}
\end{table}

%% file: tables/vlm_results.tex
\begin{table}[t]
\centering
\caption{
SOCO evaluation results for LVLMs.
All settings show the target image with candidate keypoints; only the query differs.
\textit{Vis.} uses a marked source image, \textit{Vis.+Desc.} additionally provides the keypoint description, and \textit{Desc.} uses only the keypoint description as query.
Gray values denote the absolute difference to \textit{Vis.}.
}
\label{table:vlm_res}
\vspace{-10pt}
\makebox[\linewidth][c]{%
\begin{minipage}[t]{0.68\linewidth}
\vspace{0pt}
\centering
\scriptsize
\setlength{\tabcolsep}{5pt}
\renewcommand{\arraystretch}{1.05}
\begin{tabular}{lccc}
\toprule
Method & Vis. & Vis.+Desc. & Desc. \\
\midrule
\textit{Baselines} &&&\\
Random & 0.4 & 0.4{\textcolor{gray}{\tiny +0.0}} & 0.4{\textcolor{gray}{\tiny +0.0}} \\
Random++ & 25.0 & 25.0{\textcolor{gray}{\tiny +0.0}} & 25.0{\textcolor{gray}{\tiny +0.0}} \\
DINOv2 & 81.0 & 81.0{\textcolor{gray}{\tiny +0.0}} & 81.0{\textcolor{gray}{\tiny +0.0}} \\
\midrule
\textit{LVLMs} &&&\\
LLaVA-OV-7B~\cite{lillavaov} & 2.9 & 14.1{\textcolor{gray}{\tiny +11.2}} & 24.3{\textcolor{gray}{\tiny +21.4}} \\
InternVL3.5-8B~\cite{wang2025internvl3_5} & 24.9 & 38.5{\textcolor{gray}{\tiny +13.6}} & 39.6{\textcolor{gray}{\tiny +14.7}} \\
Qwen2.5-VL-3B~\cite{Qwen2.5-VL} & 5.2 & 17.4{\textcolor{gray}{\tiny +12.2}} & 29.9{\textcolor{gray}{\tiny +24.7}} \\
Qwen2.5-VL-7B~\cite{Qwen2.5-VL} & 19.4 & 30.8{\textcolor{gray}{\tiny +11.4}} & 39.1{\textcolor{gray}{\tiny +19.7}} \\
Qwen3-VL-4B~\cite{bai2025qwen3vl} & 8.6 & 18.0{\textcolor{gray}{\tiny +9.4}} & 44.4{\textcolor{gray}{\tiny +35.8}} \\
Qwen3-VL-8B~\cite{bai2025qwen3vl} & 34.2 & 30.8{\textcolor{gray}{\tiny $-$3.4}} & 54.0{\textcolor{gray}{\tiny +19.8}} \\
GPT4o~\cite{hurst2024gpt} & 30.2 & 30.9{\textcolor{gray}{\tiny +0.7}} & 37.6{\textcolor{gray}{\tiny +7.4}} \\
\bottomrule
\end{tabular}%

\end{minipage}%
\hspace{0.01\linewidth}%
\begin{minipage}[t]{0.32\linewidth}
\vspace{10pt}
{\scriptsize LVLM evaluation settings.\vspace{5pt}}
\centering
\includegraphics[width=\linewidth]{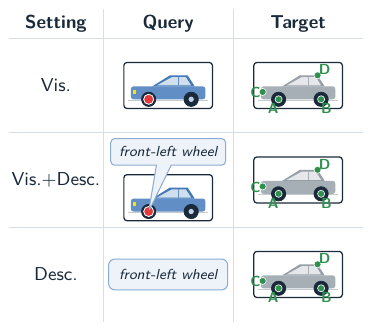}

\end{minipage}%
}
\vspace{-10pt}
\end{table}

%% file: figures/cor_with_r.tex
\begin{tikzpicture}
\begin{axis}[
  width=\linewidth,
  height=4cm,
  ymin=-0.15,
  ymax=1.20,
  ylabel={Pearson $r$ $\pm$ 95\% CI},
  xmin=0.45,
  xmax=7.55,
  xtick={1,2,3,4,5,6,7},
  xticklabels={Seg,Det3,Pose,MVC,Track,Norm,Dep},
  xticklabel style={rotate=25,anchor=north,font=\tiny,yshift=-0.15em},
  ylabel near ticks,
  yticklabel style={font=\tiny,xshift=0.25em},
  ylabel style={font=\tiny,yshift=-0.4em},
  legend style={
    at={(0.5,1.0)},
    anchor=south,
    draw=none,
    fill=none,
    font=\tiny,
    legend columns=2,
    cells={anchor=base west},
    legend image post style={yshift=0.15em},
    /tikz/every even column/.append style={column sep=0.8em},
  },
  axis line style={black!55},
  tick style={black!55},
  grid=major,
  grid style={black!8},
  error bars/y dir=both,
  error bars/y explicit,
]
\addplot[black!45,densely dashed,forget plot] coordinates {(0.45,0) (7.55,0)};

\addplot+[
  xshift=-1.0pt,
  only marks,
  mark=*,
  mark size=1.8pt,
  green!55!black,
  mark options={draw=green!55!black,fill=green!55!black},
  error bars/error bar style={green!55!black, line width=0.45pt},
] table[x=x,y=y,y error plus=errp,y error minus=errm] {
x y errm errp
0.90 0.629 0.207 0.166
1.90 0.892 0.091 0.057
2.90 0.692 0.164 0.131
3.90 0.943 0.031 0.026
4.90 0.907 0.063 0.049
5.90 0.737 0.064 0.130
6.90 0.798 0.130 0.090
};
\addlegendentry{SOCO}

\addplot+[
  xshift=1.0pt,
  only marks,
  mark=square*,
  mark size=1.7pt,
  purple!75!black,
  mark options={draw=purple!75!black,fill=purple!75!black},
  error bars/error bar style={purple!75!black, line width=0.45pt},
] table[x=x,y=y,y error plus=errp,y error minus=errm] {
x y errm errp
1.10 0.399 0.278 0.247
2.10 0.393 0.366 0.238
3.10 0.348 0.277 0.218
4.10 0.266 0.353 0.296
5.10 0.286 0.392 0.308
6.10 0.395 0.311 0.303
7.10 0.298 0.229 0.272
};
\addlegendentry{kNN}

\end{axis}
\end{tikzpicture}

%% file: figures/delta_cor.tex
\begin{tikzpicture}
\begin{axis}[
  width=\linewidth,
  height=4cm,
  ymin=-0.15,
  ymax=1.20,
  ylabel={$\Delta r$ (SOC -- kNN)},
  symbolic x coords={Seg,Det3,Pose,MVC,Track,Norm,Dep},
  xtick=data,
  xticklabel style={rotate=25,anchor=north,font=\tiny,yshift=-0.15em},
  ylabel near ticks,
  yticklabel style={font=\tiny,xshift=0.25em},
  ylabel style={font=\tiny,yshift=-0.4em},
  legend style={
    at={(0.5,1.0)},
    anchor=south,
    draw=none,
    fill=none,
    font=\tiny,
    legend columns=2,
    cells={anchor=base west},
    legend image post style={yshift=0.15em},
    /tikz/every even column/.append style={column sep=0.8em},
  },
  axis line style={black!55},
  tick style={black!55},
  grid=major,
  grid style={black!8},
  error bars/y dir=both,
  error bars/y explicit,
]
\addplot[black!45,densely dashed,forget plot] coordinates {(Seg,0) (Dep,0)};

\addplot+[
  xshift=-1.0pt,
  only marks,
  mark=*,
  mark size=1.8pt,
  blue!70!black,
  error bars/error bar style={blue!70!black, line width=0.45pt},
] table[x=metric,y=delta,y error plus=errp,y error minus=errm] {
metric delta errm errp
Seg  0.246 0.229 0.225
Pose    0.293 0.273 0.296
MVC  0.679 0.287 0.344
Track   0.625 0.287 0.363
Norm 0.352 0.279 0.540
Dep   0.520 0.234 0.298
Det3   0.536 0.222 0.359
};
\addlegendentry{All models}

\addplot+[
  xshift=1.0pt,
  only marks,
  mark=square*,
  mark size=1.7pt,
  orange!85!black,
  error bars/error bar style={orange!85!black, line width=0.45pt},
] table[x=metric,y=delta,y error plus=errp,y error minus=errm] {
metric delta errm errp
Seg  0.434 0.400 0.394
Pose    0.356 0.333 0.442
MVC  0.378 0.272 0.349
Track   0.514 0.359 0.339
Norm 0.758 0.425 0.350
Dep   0.814 0.414 0.331
Det3   0.691 0.416 0.370
};
\addlegendentry{Dense SSL}

\end{axis}
\end{tikzpicture}

%% file: sec/5_conclusion.tex
\section{Conclusion}

We introduced Semantic Object Correspondence (SOC), a principled formulation of semantic correspondence that explicitly models the relationship between object parts and the overall object structure, providing a clearer separation between geometric matching and semantic object-level understanding.
Building on this formulation, we developed SOCO, a large-scale benchmark that provides hierarchical part annotations, cross-category correspondences, and accompanying language descriptions, thus addressing the core limitations of existing datasets.

Through extensive evaluation of vision and multimodal foundation models, we demonstrated that SOCO exposes differences in their ability to capture fine-grained, object-centric structure.
The taxonomy makes three failure modes separately measurable: the CC$\to$SOC gap isolates repeated-part disambiguation, the SOC$\to$Cross-SOC gap isolates category-specific concept encoding, and the Vis.\,vs.\,Desc.\ gap in LVLMs separates cross-image visual matching from language-grounded part localization.
Our results show that:
(1) models reliably match semantic concepts but struggle with object-level geometry;
(2) cross-category correspondence remains challenging even for the strongest vision backbones;
(3) large vision–language models are stronger at text-prompted keypoint localization than at visual-reference correspondence, revealing a gap between language-grounded localization and visual matching; and
(4) SOC performance correlates with dense vision tasks more strongly than ImageNet $k$NN, making SOC a powerful zero-shot diagnostic for representation quality.

SOCO provides a unified testbed for analyzing structured part-level visual and multimodal understanding in modern foundation models.  
We hope it serves as a stepping stone toward models that not only recognize objects but also understand their parts and structural relationships in a way that generalizes across categories and modalities.

%% file: sec/X_ack.tex
\section*{Acknowledgments}
AK acknowledges support via his Emmy Noether Research Group funded by the German Research Foundation (DFG) under grant number 468670075.
We thank Matthis Heimberg for early analyses and experiments.

%% file: sec/X_suppl.tex
\clearpage
\setcounter{page}{1}
 {
   \newpage
        \centering
        \Large
        \textbf{\papertitle}\\
        \vspace{0.5em}Supplementary Material \\
        \vspace{1.5em}
   }

\appendix
\def\theHpage{appendix.\arabic{page}}
\def\theHsection{appendix.\Alph{section}}
\def\theHsubsection{appendix.\Alph{section}.\arabic{subsection}}
\def\theHsubsubsection{appendix.\Alph{section}.\arabic{subsection}.\arabic{subsubsection}}

\renewcommand{\thetable}{r\arabic{table}}
\renewcommand{\thefigure}{r\arabic{figure}}
\setcounter{table}{0}
\setcounter{figure}{0}

\noindent To complement the main paper, this supplementary material provides more experimental results and implementation details.

\addcontentsline{toc}{part}{Appendix}

\medskip
\noindent\textbf{Outline of Supplementary Material}
\par
\begingroup
\footnotesize
\providecommand{\authcount}[1]{}
\etocsetlevel{title}{6}
\etocsetlevel{author}{6}
\etocinline
\etocsettocstyle{}{}
\etocsetstyle{part}{}{}{}{}
\etocsetstyle{section}
  {\par\addvspace{0.45em}}
  {}
  {%
    \noindent
    \makebox[3.2em][l]{\etocifnumbered{\textbf{\etocnumber}}{}}%
    \textbf{\etocname}\nobreak\leaders\hbox to 0.6em{\hss.\hss}\hfill
    \textbf{\etocpage}\par
  }
  {}
\etocsetstyle{subsection}
  {}
  {}
  {%
    \noindent\hspace*{1.2em}%
    \makebox[4.2em][l]{\etocifnumbered{\etocnumber}{}}%
    \etocname\nobreak\leaders\hbox to 0.6em{\hss.\hss}\hfill
    \etocpage\par
  }
  {}
\etocsetstyle{subsubsection}{}{}{}{}
\etocsetnexttocdepth{2}
\localtableofcontents*
\endgroup

\section{More Quantitative Results on SOCO}

This section will present additional results on the SOCO dataset: 
\Cref{sec:soco_subsets} and \Cref{sec:various_pck} present model evaluations on various subsets and PCK levels.
\Cref{sec:soco_eval_more} includes an evaluation of further models in addition to the results presented in the main paper.
\Cref{sec:soc-per-category} reports per-category results.

\subsection{Evaluation on Supercategories} \label{sec:soco_subsets}
Complementing the per-supercategory SOC results in the main paper (\cref{tab:soc_performance}), \cref{tab:soco7_soc_subset_perkpt} reports both SOC and CC for each of the four super-categories \textit{transportation}, \textit{hand-held}, \textit{furniture}, and \textit{animals}, so the CC$\to$SOC gap per super-category can be read off directly.
Interestingly, models perform worst for the furniture super-category for SOC but CC performance is even better than for the other super-categories.
These larger drops might be attributed to the fact that furniture objects have more object parts that have locally similar semantics, such as the legs of a chair.
Similarly, drops are large as well for transportation categories, as they contain repeated object parts, e.g., wheels.
For the SOC setting of the furniture categories, DINOv3 clearly outperforms DINOv2, indicating that DINOv3 captures geometric position better. 
This trend is similar for the transportation category where drops from CC to SOC are smaller for DINOv3 than for DINOv2.

\subsection{Complementary Evaluation Protocols}
To supplement the nearest neighbor strategy as reported in the main paper, we present more additional evaluation strategies in \cref{tab:soco7_linear_soft_geo}.

\noindent 1) First, we perform an evaluation based on window softargmax~\cite{zhang2023tale,zhang2024telling} (\textit{soft-eval}). 
This consistently improves results but not by a large margin.

\noindent 2) Second, we train a linear probe (shared across all patches) supervised and evaluated, each on 100 pairs of disjunct images (\textit{SOC linear}), using an image resolution of 500px.
The performance improves substantially across all models.
This shows that selecting a subspace of the dense features results in a learned manner leads to better matching performance, as information is discarded that changes across instances and a positional bias is added.
While the best models consistently remain the best, some model rankings substantially change. For example, SD clearly improves.

\noindent 3) Third, we evaluate SOC using a fixed number of $60\cdot 60 = 3600$ patches (instead of a fixed image resolution). 
This benefits models with a larger patch size, eg, DINOv3 or C-RADIO. 
Still, rankings do not substantially change.

\subsection{Evaluation of Geometric Awareness (SOC-geo)}\label{suppl:sec:socgeo}
Relying on the explicit separation of geometric attributes and semantic concept, we further evaluate \textit{SOC-geo}:
This evaluates specifically whether a models is capable of differentiating the geometric positions of keypoints of the same concept.
Given one source keypoint, the argmax is computed over all instances of the same concept for a target image of the same category.
We only select image pairs where there are at least two pairs to match, which results in around $100k$ evaluated keypoint pairs.
The random performance is 41.24\% for this setting: The number of evaluated keypoints varies across categories and images. E.g., a car wheel might appear two or three times on an image but for a chair all four legs are often visible.
We present the results in \cref{tab:soco7_linear_soft_geo}
Here, the model rankings change clearly:
For example, DINOv3 outperforms DINOv2 and SD is performing best, indicating that it encodes object part position more effectively.
This is in line with the analysis of the gap between SOC and CC for various supercategories.

\subsection{Evaluation for Varying PCK Thresholds}\label{sec:various_pck}
\Cref{tab:soco7_soc_perkpt_perimg} presents results for various PCK thresholds with pair averaging (\textit{per-img}).
The performances substantially drop for smaller thresholds.
Interestingly, results drop less for SD than for other models.

\begin{table*}[t]
\caption{
\textbf{Model performances on SOCO across supercategories}.
The results are presented in the format (SOC $|$ CC) for the four supercategories of SOCO.
The model performances heavily vary for different categories and the gaps between SOC and CC change across categories and models.
}
\label{tab:soco7_soc_subset_perkpt}
\centering
\footnotesize
\setlength{\tabcolsep}{5pt}
\begin{tabular}{lcccc}
\toprule
Model & Transportation & Hand-held & Furniture & Animals  \\
\midrule
DINOv1 & 29.1 $|$ 43.0 & 32.9 $|$ 43.2 & 27.4 $|$ 49.5 & 31.4 $|$ 43.3  \\
DINOv2 & \textbf{56.9} $|$ 76.4 & \textbf{61.6} $|$ 74.9 & 45.5 $|$ 77.5 & \textbf{66.3} $|$ 83.1  \\
DINOv3 & 51.6 $|$ 66.1 & 57.4 $|$ 66.3 & \textbf{59.9} $|$ 76.2 & 56.6 $|$ 72.5  \\
iBOT & 36.1 $|$ 52.3 & 40.1 $|$ 51.4 & 32.8 $|$ 59.8 & 43.9 $|$ 58.0  \\
I-JEPA & 41.5 $|$ 57.1 & 51.0 $|$ 58.2 & 46.7 $|$ 71.7 & 47.7 $|$ 61.4  \\
C-RADIOv3 & 51.7 $|$ 69.9 & 48.1 $|$ 62.9 & 39.7 $|$ 68.5 & 54.8 $|$ 71.4  \\
DUNE & 40.0 $|$ 55.6 & 50.5 $|$ 59.3 & 51.0 $|$ 72.0 & 46.6 $|$ 61.0  \\
SD 2.1 & 46.4 $|$ 60.5 & 47.0 $|$ 54.9 & 50.9 $|$ 66.6 & 50.9 $|$ 61.9  \\
CroCov2 & 11.6 $|$ 17.5 & 12.3 $|$ 18.1 & 10.2 $|$ 18.2 & 8.1 $|$ 11.2  \\
MAE & 10.0 $|$ 15.9 & 11.9 $|$ 17.5 & 11.7 $|$ 20.6 & 7.1 $|$ 10.3  \\
PIXIO & 37.3 $|$ 50.4 & 40.2 $|$ 48.6 & 46.7 $|$ 65.3 & 34.2 $|$ 45.2  \\
CLIP & 17.9 $|$ 27.8 & 14.7 $|$ 23.1 & 11.7 $|$ 26.1 & 16.8 $|$ 24.1  \\
PE-Spatial & 46.5 $|$ 64.6 & 43.4 $|$ 56.6 & 38.7 $|$ 64.9 & 48.4 $|$ 65.0  \\
QWEN-L & 28.1 $|$ 39.4 & 28.4 $|$ 37.2 & 29.3 $|$ 45.0 & 22.5 $|$ 32.4  \\
\bottomrule
\end{tabular}
\end{table*}

\begin{table*}[t]
\caption{
\textbf{SOCO results with various evaluation protocols.}
We report evaluation with window soft argmax (\textit{soft-eval}), trained with a linear probe, and with a fixed number of patches per image (\textit{SOC-fixed}).
}
\centering
\footnotesize
\setlength{\tabcolsep}{5pt}
\begin{tabular}{lccc}
\toprule
Model & SOC soft-eval & SOC-fixed &SOC linear \\
\midrule
DINOv1 & 32.3 & 29.3 &42.2 \\
DINOv2 & \textbf{62.7} & \textbf{60.6} &\textbf{69.8} \\
DINOv3 & 55.7 & 56.4 &69.2 \\
iBOT & 41.4 & 38.1 &50.7 \\
I-JEPA & 47.7 & 47.2 &61.6 \\
C-RADIOv3 & 52.4 & 52.3 &62.1 \\
DUNE & 48.0 & 46.5 &58.1 \\
SD 2.1 & 49.9 & 48.8 &54.4 \\
CroCov2 & 10.2 & 9.1 &29.1 \\
MAE & 9.1 & 9.3 &29.9 \\
PIXIO & 35.7 & 33.2 &63.3 \\
CLIP & 17.0 & 14.8 &35.1 \\
PE-Spatial & 47.8 & 44.0 &55.3 \\
QWEN-L & 26.3 & 23.3 & 34.31 \\
\bottomrule
\end{tabular}
\label{tab:soco7_linear_soft_geo}
\end{table*}

\begin{table*}[t]
\caption{\textbf{Model performances on SOCO across multiple thresholds}.
The results are presented in the format (SOC $|$ CC) for both pair averaging and per-keypoint reduction.
}
\label{tab:soco7_soc_perkpt_perimg}
\centering
\tiny
\setlength{\tabcolsep}{4pt}
\begin{tabular}{l ccc ccc}
\toprule
& \multicolumn{3}{c}{\textbf{Pair Averaging}} & \multicolumn{3}{c}{\textbf{Per Keypoint}} \\
\cmidrule(lr){2-4}\cmidrule(lr){5-7}
Model & PCK@0.10 & PCK@0.05 & PCK@0.02 & PCK@0.10 & PCK@0.05 & PCK@0.02 \\
\midrule
DINOv1 & 30.6 $|$ 43.8 & 16.0 $|$ 24.0 & 4.6 $|$ 7.0 & 32.1 $|$ 44.8 & 16.9 $|$ 24.8 & 4.9 $|$ 7.2 \\
DINOv2 & \textbf{60.4} $|$ 78.9 & \textbf{41.6} $|$ 57.2 & \textbf{15.4} $|$ 21.5 & \textbf{61.7} $|$ 79.6 & \textbf{43.0} $|$ 58.2 & \textbf{16.1} $|$ 22.0 \\
DINOv3 & 55.5 $|$ 69.7 & 36.3 $|$ 47.2 & 12.2 $|$ 15.7 & 57.8 $|$ 70.4 & 38.3 $|$ 48.0 & 13.0 $|$ 16.1 \\
iBOT & 39.6 $|$ 55.2 & 21.0 $|$ 30.9 & 5.8 $|$ 8.6 & 41.4 $|$ 56.3 & 22.2 $|$ 31.8 & 6.2 $|$ 8.9 \\
I-JEPA & 46.3 $|$ 60.5 & 31.5 $|$ 42.8 & 11.0 $|$ 15.1 & 48.0 $|$ 61.3 & 33.0 $|$ 43.7 & 11.6 $|$ 15.4 \\
C-RADIOv3 & 51.1 $|$ 69.0 & 31.4 $|$ 44.5 & 9.8 $|$ 13.9 & 52.4 $|$ 70.0 & 32.6 $|$ 45.6 & 10.2 $|$ 14.3 \\
DUNE & 45.7 $|$ 60.1 & 30.2 $|$ 41.4 & 11.1 $|$ 15.2 & 48.0 $|$ 61.0 & 32.0 $|$ 42.3 & 11.9 $|$ 15.7 \\
SD 2.1 & 48.8 $|$ 60.6 & 35.4 $|$ 44.8 & 14.6 $|$ 18.2 & 52.2 $|$ 63.0 & 38.3 $|$ 46.8 & 15.8 $|$ 18.8 \\
CroCov2 & 10.2 $|$ 15.1 & 4.8 $|$ 7.2 & 1.4 $|$ 2.0 & 10.9 $|$ 15.4 & 5.2 $|$ 7.4 & 1.5 $|$ 2.1 \\
MAE & 9.4 $|$ 14.4 & 3.9 $|$ 6.0 & 1.0 $|$ 1.5 & 10.0 $|$ 14.7 & 4.2 $|$ 6.2 & 1.1 $|$ 1.6 \\
PIXIO & 37.5 $|$ 49.5 & 20.5 $|$ 28.2 & 6.0 $|$ 8.3 & 39.8 $|$ 50.5 & 22.0 $|$ 28.9 & 6.5 $|$ 8.5 \\
CLIP & 16.2 $|$ 25.2 & 7.2 $|$ 11.8 & 1.8 $|$ 2.9 & 17.1 $|$ 26.1 & 7.7 $|$ 12.2 & 1.9 $|$ 3.1 \\
PE-Spatial & 45.9 $|$ 63.2 & 29.6 $|$ 43.1 & 10.0 $|$ 14.7 & 47.1 $|$ 64.1 & 30.7 $|$ 44.1 & 10.5 $|$ 15.0 \\
QWEN-L & 26.0 $|$ 36.8 & 13.0 $|$ 19.1 & 2.7 $|$ 4.0 & 27.7 $|$ 37.4 & 14.0 $|$ 19.5 & 3.0 $|$ 4.1 \\
\bottomrule
\end{tabular}
\end{table*}

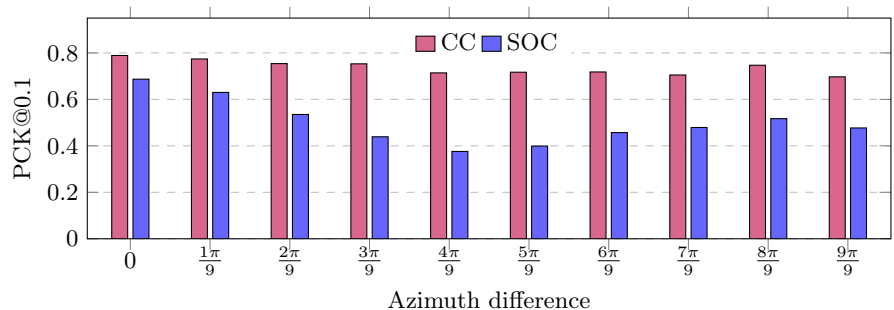
\begin{figure}[ht]
  \centering
\begin{tikzpicture}
\begin{axis}[
    ybar,
    bar width=6pt,
    width=\linewidth,
    height=4.5cm,
    ymin=0, ymax=0.95,
    ylabel={PCK@0.1},
    symbolic x coords={b1,b2,b3,b4,b5,b6,b7,b8,b9,b10},
    xtick={b1,b2,b3,b4,b5,b6,b7,b8,b9,b10},
    xticklabels={
        $0$,
        $\frac{1\pi}{9}$,
        $\frac{2\pi}{9}$,
        $\frac{3\pi}{9}$,
        $\frac{4\pi}{9}$,
        $\frac{5\pi}{9}$,
        $\frac{6\pi}{9}$,
        $\frac{7\pi}{9}$,
        $\frac{8\pi}{9}$,
        $\frac{9\pi}{9}$
    },
    x tick label style={yshift=-4pt, rotate=0, anchor=center},
    xlabel={Azimuth difference},
    enlarge x limits=0.06,
    legend style={
        at={(0.5,0.98)},
        anchor=north,
        legend columns=-1,
        fill=white,
        fill opacity=1.,
        draw=none
    },
    legend image code/.code={
        \draw[#1,fill=#1,mark=none] (0cm,-0.1cm) rectangle (0.25cm,0.1cm);
    },
    ymajorgrids=true,
    grid style=dashed,
]

\addplot[fill=purple!60] coordinates {
    (b1,0.789)
    (b2,0.774)
    (b3,0.754)
    (b4,0.753)
    (b5,0.714)
    (b6,0.717)
    (b7,0.718)
    (b8,0.705)
    (b9,0.747)
    (b10,0.697)
};

\addplot[fill=blue!60] coordinates {
    (b1,0.687)
    (b2,0.630)
    (b3,0.535)
    (b4,0.439)
    (b5,0.376)
    (b6,0.399)
    (b7,0.457)
    (b8,0.479)
    (b9,0.517)
    (b10,0.477)
};
\legend{CC, SOC}
\end{axis}
\end{tikzpicture}
  \caption{\textbf{PCK of DINOv2/b for increasing azimuth variation} between two images, averaged over all categories.
  While the concept correspondence (CC) remains stable for larger viewpoint changes, SOC performance drops with a minimum for a relative orientation of objects of $\pi/2$.
  }
  \label{fig:delta_azimuth_bins}
\end{figure}

\subsection{Analysis of Viewpoint Variation}
  
\Cref{fig:delta_azimuth_bins} presents the performance variation for varying viewpoint differences between the two matched objects.
We exemplarily report the results for DINOv2, here.
For this, we extract the labeled 3D pose as given by \cite{ma2024imagenet3d}, compute the difference between the azimuth angles, and bin those differences.
Subsequently, we compute the average performance of all matches within the considered bin.
The SOC performance is lowest for a $\pi/2$ viewpoint difference, as this is the most challenging scenario, as objects are rotated by 90$^\circ$ and object parts are harder to disambiguate.
For larger viewpoint changes, there are fewer ambiguous keypoints, increasing the performance again. For example, when two cars are observed from the left and the right side, there are not co-visible wheels that are to be matched.
At the same time, CC performance remains comparably constant, indicating the pure semantic matching is still effective but geometric differentiation is limited when objects are not in the same pose.

\subsection{Evaluation of More VFMs}\label{sec:soco_eval_more}
In addition to the models presented in the main paper, we evaluate additional models on the SOCO dataset and we present the results in \Cref{suppl:tab:soc_performance_more_models}.
We find that larger models typically outperform the base models that are evaluated in the main paper, e.g., the large variants of DINOv2, DINOv3, or C-RADIO v4.
DINOv2, DINOv3, and C-RADIOv4 reach comparable performance on SOC.
Additional results including the current SOTA-models on SPair-71k are presented in \Cref{suppl:tab:soc_performance_more_models_cleandift}, following the implementation of CleanDIFT~\cite{stracke2025cleandift} and GeoAware-SC~\cite{zhang2024telling}.
Here, we also evaluate SOC for weakly-supervised models:
DIY-SC~\cite{dunkel2025yourself} and SD+DINO~\cite{zhang2024telling} with CLIP embeddings fine-tuned on panoptic segmentation.
Furthermore, we also evaluate the supervised variant of \cite{zhang2024telling} relying on SD+DINO features and only relying on DINO features. We train both from scratch on SPair-71k.
While weak supervision specifically used to improve semantic correspondence improves performance on this dataset as well, the performance of the models trained supervised on SPair-71k clearly drops compared to the SPair-71k test dataset performance that is substantially larger than 80\%.
This indicates that SOCO captures new categories that are different to the SPair-71k categories.
Particularly, while animal, transportation, and furniture categories clearly improve compared to the zero-shot approach with SD+DINO features, the matching performance drops by 8.61 points for hand-held objects.

\begin{table}[tbp]
\begin{scriptsize}
\centering
\caption{
\textbf{Evaluation of additional models on SOCO} following the implementation of GeoAware-SC and CleanDIFT~\cite{stracke2025cleandift,zhang2024telling,mariotti2024improving,dunkel2025yourself}.
Results are reported for PCK@0.1 (\textit{per-img}).
}
\renewcommand{\arraystretch}{1.1}
\setlength{\tabcolsep}{6pt}
\label{suppl:tab:soc_performance_more_models_cleandift}

\begin{tabular}{lc}
  \toprule
  \textbf{Model} & \textbf{SOC} \\
  \midrule
  SD (ft. for pan. seg., Geo-SC)& 47.7\\
  SD + DINO (SD ft pan. seg., Geo-SC)& 62.9\\
  CleanDIFT + DINO & 63.4\\
  DIY-SC (weakl. sup.)& 69.2 \\
  \midrule
  TLR (w SD+DINO) (sup. SPair-71k)& 72.9\\
  TLR (w DINO) (sup. SPair-71k)& 72.7\\
  \bottomrule
  \end{tabular}

\end{scriptsize}

\end{table}

\begin{table*}[ht]
\caption{
\textbf{Model performances on SOCO} across concept correspondence (CC), semantic object correspondence (SOC) and its cross-category variants (Cross-SOC).
This table extends the table presented in the main paper.
}
\scriptsize
\renewcommand{\arraystretch}{1.0}
\setlength{\tabcolsep}{2pt}
\centering
\begin{minipage}[t]{0.48\textwidth}\centering
\begin{tabular}{lccc}
\toprule
Model & CC & SOC & Cross \\
\midrule
C-RADIO-v3-B & 69.0 & 51.1 & 46.3 \\
C-RADIO-v3-L & 76.8 & 58.8 & 53.8 \\
C-RADIO-v4-SO400M & 79.6 & 61.5 & 56.3 \\
CLIP-B/16 (DataComp) & 35.1 & 23.3 & 16.5 \\
CLIP-B/16 (LAION2B) & 26.3 & 16.7 & 11.7 \\
CLIP-B/16 (OpenAI) & 25.2 & 16.2 & 11.4 \\
CLIP-L/14 (DataComp) & 46.5 & 32.1 & 24.6 \\
CLIP-L/14 (LAION2B) & 36.2 & 24.4 & 17.6 \\
CLIP-L/14 (OpenAI) & 37.9 & 25.9 & 18.7 \\
ConvNeXt-B (FCMAE) & 49.6 & 34.6 & 18.4 \\
ConvNeXt-B (IN22K) & 34.2 & 23.5 & 10.4 \\
ConvNeXt-B (OpenCLIP) & 34.1 & 24.3 & 15.5 \\
ConvNeXt-B (OpenCLIP-AR) & 44.0 & 31.1 & 22.2 \\
CroCov2-B/16 & 15.1 & 10.2 & 7.8 \\
DeiT3-B/16 & 35.5 & 23.3 & 12.1 \\
DeiT3-L/16 & 47.5 & 32.6 & 17.1 \\
DINO-B/16 & 43.8 & 30.6 & 23.9 \\
DINO-S/16 & 43.1 & 30.0 & 23.2 \\
DINOv2-B/14 & 78.9 & 60.4 & 54.9 \\
DINOv2-B/14-reg & 76.3 & 58.0 & 50.6 \\
DINOv2-L/14 & 80.7 & 62.4 & 56.8 \\
DINOv2-S/14 & 74.0 & 56.7 & 50.0 \\
DINOv3-B/16 & 69.7 & 55.5 & 49.4 \\
DINOv3-ConvNeXt-B & 60.9 & 47.6 & 39.1 \\
DINOv3-ConvNeXt-L & 62.9 & 48.9 & 41.8 \\
DINOv3-ConvNeXt-S & 61.0 & 47.9 & 37.5 \\
DINOv3-ConvNeXt-T & 60.4 & 46.8 & 36.3 \\
DINOv3-L/16 (LVD) & 76.6 & 61.4 & 56.2 \\
DINOv3-L/16 (SAT) & 29.4 & 21.1 & 16.1 \\
DINOv3-S/16 & 65.6 & 51.5 & 45.2 \\
DINOv3-S+/16 & 69.3 & 55.2 & 49.4 \\
DUNE-B/14 & 57.7 & 44.0 & 37.8 \\
DUNE-B/14 (336px) & 56.1 & 42.5 & 36.4 \\
DUNE-B/14 (448px) & 60.1 & 45.7 & 39.4 \\
DUNE-S/14 & 50.7 & 38.2 & 32.5 \\
I-JEPA-H/16 & 60.5 & 46.3 & 38.4 \\
iBOT-B/16 & 55.2 & 39.6 & 34.1 \\
iBOT-B/16 (IN22K) & 53.1 & 38.4 & 32.9 \\
iBOT-L/16 & 64.1 & 47.4 & 42.8 \\
iBOT-L/16 (IN22K) & 66.5 & 49.4 & 44.8 \\
iBOT-S/16 & 48.2 & 33.8 & 27.9 \\
MAE-B/16 & 14.4 & 9.4 & 7.1 \\
MAE-L/16 & 13.7 & 8.8 & 6.7 \\
MetaCLIP2-B/16 & 31.2 & 20.2 & 15.4 \\
MetaCLIP2-L/14 & 38.7 & 25.9 & 18.3 \\
MetaCLIP2-S/16 & 24.9 & 15.8 & 12.1 \\
MiDaS-L/16 & 40.4 & 27.1 & 21.4 \\
PE-Spatial-B/16 & 63.2 & 45.9 & 40.9 \\
PIXIO-B/16 & 49.5 & 37.5 & 32.9 \\
PIXIO-L/16 & 51.3 & 38.6 & 32.9 \\
Qwen2.5-VL-7B & 36.8 & 26.0 & 22.8 \\
SAM-B & 35.2 & 25.5 & 20.4 \\
SAM-L & 35.9 & 26.0 & 20.6 \\
SD 2.1 & 60.5 & 49.0 & 42.7 \\
SigLIP-B/16 & 14.1 & 8.8 & 7.0 \\
SigLIP-L/16 & 14.5 & 9.6 & 7.5 \\
V-JEPA2-L & 42.5 & 31.2 & 26.7 \\
V-JEPA2.1-B & 60.4 & 44.5 & 38.6 \\
V-JEPA2.1-L & 69.7 & 53.2 & 48.1 \\
VGGT-1B & 21.2 & 15.8 & 12.5 \\
\bottomrule
\end{tabular}
\end{minipage}
\label{suppl:tab:soc_performance_more_models}
\end{table*}

\subsection{Category-Specific Results}\label{sec:soc-per-category}
\Cref{suppl:tab:soc_performance_per_cat} presents per-category results for DINOv2-B at varying PCK thresholds.
The gaps between SOC and CC largely depend on the considered category.
Similarly, reducing the threshold for the PCK calculation has a varying effect on different categories.

\begin{table*}[ht]
\caption{
\textbf{SOCO per-category results} evaluated with DINOv2-B across multiple PCK thresholds (pair averaging).
Each cell shows CC $|$ SOC.
}
\tiny
\renewcommand{\arraystretch}{1.0}
\setlength{\tabcolsep}{1pt}
\centering
\begin{minipage}[t]{0.48\textwidth}\centering
\begin{tabular}{lccc}
\toprule
& @0.10 & @0.05 & @0.02 \\
\midrule
aeroplane & 83.4 $|$ 72.9 & 69.7 $|$ 59.1 & 34.3 $|$ 28.5 \\
ambulance & 87.7 $|$ 54.1 & 74.0 $|$ 43.6 & 36.4 $|$ 20.8 \\
american black bear & 72.7 $|$ 54.0 & 44.3 $|$ 32.3 & 12.8 $|$ 8.9 \\
arctic fox & 81.9 $|$ 65.4 & 56.8 $|$ 44.0 & 17.3 $|$ 13.1 \\
armchair & 79.2 $|$ 48.6 & 57.7 $|$ 35.9 & 21.1 $|$ 13.1 \\
ax & 81.2 $|$ 55.5 & 60.8 $|$ 35.5 & 27.0 $|$ 16.5 \\
bed & 74.8 $|$ 40.5 & 63.7 $|$ 34.6 & 32.3 $|$ 17.6 \\
bench & 79.1 $|$ 37.0 & 58.9 $|$ 27.5 & 22.5 $|$ 9.5 \\
bicycle & 85.7 $|$ 83.8 & 65.6 $|$ 64.6 & 26.1 $|$ 25.6 \\
bicycle pump & 89.7 $|$ 69.1 & 80.8 $|$ 57.1 & 43.0 $|$ 28.5 \\
bighorn & 85.1 $|$ 67.6 & 55.6 $|$ 40.8 & 15.4 $|$ 12.5 \\
boston bull & 87.8 $|$ 59.5 & 62.8 $|$ 40.5 & 20.0 $|$ 13.1 \\
brittany spaniel & 89.4 $|$ 74.3 & 62.1 $|$ 50.1 & 21.3 $|$ 17.3 \\
brown bear & 75.7 $|$ 61.2 & 47.0 $|$ 36.5 & 14.5 $|$ 11.2 \\
bullet train & 62.4 $|$ 48.7 & 38.3 $|$ 28.8 & 11.6 $|$ 8.5 \\
bus & 82.6 $|$ 54.4 & 68.0 $|$ 43.9 & 29.8 $|$ 18.4 \\
cabin tractor & 72.8 $|$ 54.9 & 48.4 $|$ 34.7 & 11.5 $|$ 8.1 \\
cairn & 78.6 $|$ 62.5 & 55.7 $|$ 40.4 & 20.1 $|$ 13.7 \\
car & 90.1 $|$ 62.1 & 78.6 $|$ 54.2 & 36.0 $|$ 26.1 \\
cart & 53.8 $|$ 40.8 & 36.6 $|$ 26.6 & 14.0 $|$ 9.4 \\
chair & 85.0 $|$ 41.9 & 69.5 $|$ 33.2 & 31.9 $|$ 14.7 \\
cheetah & 89.5 $|$ 78.0 & 70.6 $|$ 59.5 & 27.8 $|$ 24.5 \\
chow & 90.1 $|$ 64.1 & 65.7 $|$ 45.1 & 23.7 $|$ 16.6 \\
cougar & 85.6 $|$ 64.1 & 60.8 $|$ 42.2 & 21.3 $|$ 15.3 \\
dishwasher & 68.4 $|$ 43.1 & 48.3 $|$ 29.1 & 17.1 $|$ 10.3 \\
dune buggy & 64.5 $|$ 45.3 & 50.4 $|$ 34.0 & 22.7 $|$ 14.5 \\
egyptian cat & 71.8 $|$ 53.1 & 44.8 $|$ 32.0 & 11.4 $|$ 8.4 \\
english springer & 75.1 $|$ 62.2 & 48.0 $|$ 39.0 & 12.5 $|$ 9.5 \\
eskimo dog & 90.7 $|$ 67.3 & 66.6 $|$ 45.0 & 22.4 $|$ 15.7 \\
eyeglasses & 74.4 $|$ 55.1 & 59.4 $|$ 43.2 & 28.2 $|$ 19.6 \\
f1 car & 77.4 $|$ 54.3 & 58.1 $|$ 37.7 & 22.9 $|$ 14.4 \\
fighter jet & 67.8 $|$ 55.0 & 54.1 $|$ 44.8 & 21.6 $|$ 18.4 \\
fire truck & 80.3 $|$ 54.5 & 55.5 $|$ 37.7 & 20.4 $|$ 12.6 \\
folding chair & 85.4 $|$ 44.6 & 71.8 $|$ 38.0 & 35.3 $|$ 19.2 \\
forklift & 79.5 $|$ 48.2 & 62.0 $|$ 34.8 & 26.4 $|$ 13.9 \\
french horn & 62.8 $|$ 34.9 & 40.5 $|$ 19.0 & 7.9 $|$ 3.7 \\
garbage truck & 81.9 $|$ 65.0 & 66.4 $|$ 48.5 & 28.8 $|$ 20.5 \\
gazelle & 90.7 $|$ 78.6 & 65.9 $|$ 50.7 & 19.4 $|$ 14.9 \\
glider & 78.3 $|$ 66.5 & 64.6 $|$ 54.5 & 31.4 $|$ 27.0 \\
go kart & 77.0 $|$ 58.1 & 52.9 $|$ 40.7 & 18.7 $|$ 12.2 \\
golden retriever & 81.9 $|$ 63.4 & 56.5 $|$ 42.3 & 21.2 $|$ 16.1 \\
gordon setter & 82.5 $|$ 68.8 & 57.5 $|$ 44.2 & 17.3 $|$ 13.4 \\
guitar & 91.2 $|$ 79.8 & 79.3 $|$ 49.4 & 35.6 $|$ 18.0 \\
hacksaw & 69.2 $|$ 56.4 & 53.6 $|$ 43.3 & 19.8 $|$ 16.3 \\
hair dryer & 67.1 $|$ 55.4 & 47.7 $|$ 37.7 & 16.0 $|$ 12.6 \\
hartebeest & 85.1 $|$ 77.0 & 60.9 $|$ 50.6 & 16.4 $|$ 12.1 \\
highchair & 74.7 $|$ 41.5 & 51.3 $|$ 27.2 & 16.5 $|$ 7.9 \\
ibex & 84.9 $|$ 70.6 & 54.8 $|$ 42.2 & 15.7 $|$ 12.0 \\
ice bear & 82.9 $|$ 66.9 & 59.1 $|$ 42.8 & 21.2 $|$ 15.2 \\
impala & 90.0 $|$ 69.2 & 63.8 $|$ 46.7 & 21.4 $|$ 15.1 \\
\bottomrule
\end{tabular}
\end{minipage}\hspace{5pt}
\begin{minipage}[t]{0.48\textwidth}\centering
\begin{tabular}{lccc}
\toprule
& @0.10 & @0.05 & @0.02 \\
\midrule
irish water spaniel & 83.6 $|$ 71.0 & 56.7 $|$ 49.4 & 18.0 $|$ 17.3 \\
iron & 60.0 $|$ 53.2 & 44.9 $|$ 40.4 & 17.9 $|$ 16.0 \\
japanese spaniel & 78.4 $|$ 59.3 & 47.5 $|$ 36.4 & 15.1 $|$ 11.9 \\
jinrikisha & 72.2 $|$ 44.4 & 54.6 $|$ 31.8 & 22.1 $|$ 12.6 \\
kettle & 68.6 $|$ 49.5 & 38.7 $|$ 27.8 & 9.8 $|$ 7.2 \\
kettle electric & 72.0 $|$ 67.4 & 46.5 $|$ 41.8 & 16.0 $|$ 14.1 \\
knife & 87.8 $|$ 83.5 & 68.0 $|$ 54.1 & 33.8 $|$ 27.6 \\
leopard & 86.2 $|$ 73.1 & 64.7 $|$ 50.8 & 23.2 $|$ 17.8 \\
megaphone & 81.5 $|$ 71.0 & 59.4 $|$ 49.4 & 20.2 $|$ 16.8 \\
microwave & 67.3 $|$ 49.5 & 47.1 $|$ 35.0 & 17.8 $|$ 13.9 \\
motor scooter & 72.1 $|$ 65.7 & 51.9 $|$ 47.0 & 22.7 $|$ 20.1 \\
motorbike & 72.9 $|$ 74.2 & 56.7 $|$ 59.2 & 26.7 $|$ 28.3 \\
office chair & 77.0 $|$ 44.9 & 55.4 $|$ 33.3 & 20.0 $|$ 12.1 \\
ox & 78.1 $|$ 62.4 & 47.5 $|$ 35.1 & 12.7 $|$ 8.8 \\
pickup truck & 87.0 $|$ 47.8 & 74.1 $|$ 39.7 & 37.0 $|$ 20.2 \\
power drill & 75.9 $|$ 66.0 & 47.6 $|$ 41.3 & 13.6 $|$ 12.7 \\
ram & 82.0 $|$ 65.0 & 50.1 $|$ 36.1 & 15.2 $|$ 10.0 \\
redbone & 90.2 $|$ 70.4 & 71.9 $|$ 53.4 & 26.1 $|$ 19.3 \\
rifle & 75.2 $|$ 70.9 & 64.4 $|$ 59.6 & 32.4 $|$ 30.3 \\
saint bernard & 89.5 $|$ 72.7 & 61.6 $|$ 46.6 & 18.1 $|$ 13.8 \\
saluki & 85.4 $|$ 76.1 & 67.5 $|$ 54.7 & 25.6 $|$ 20.4 \\
saxophone & 72.9 $|$ 63.3 & 50.3 $|$ 39.3 & 19.2 $|$ 14.4 \\
school bus & 85.7 $|$ 53.5 & 72.7 $|$ 43.8 & 36.4 $|$ 21.0 \\
segway & 67.2 $|$ 46.3 & 38.8 $|$ 26.1 & 12.6 $|$ 7.8 \\
sewing machine & 76.3 $|$ 71.1 & 59.5 $|$ 55.1 & 25.7 $|$ 24.3 \\
sloth bear & 69.4 $|$ 52.1 & 37.1 $|$ 26.9 & 8.5 $|$ 6.3 \\
snowmobile & 74.7 $|$ 64.4 & 54.6 $|$ 45.4 & 22.9 $|$ 18.1 \\
sofa & 76.8 $|$ 52.7 & 56.6 $|$ 39.7 & 22.2 $|$ 16.3 \\
soft coated terrier & 79.0 $|$ 60.4 & 41.8 $|$ 31.4 & 12.1 $|$ 9.8 \\
sorrel & 75.6 $|$ 55.8 & 44.6 $|$ 30.1 & 9.9 $|$ 6.5 \\
sports car & 90.6 $|$ 64.7 & 73.4 $|$ 47.0 & 28.9 $|$ 18.4 \\
tank & 67.6 $|$ 56.8 & 51.4 $|$ 43.5 & 19.7 $|$ 16.0 \\
teapot & 74.8 $|$ 69.0 & 47.5 $|$ 44.9 & 18.6 $|$ 18.0 \\
tibetan terrier & 68.4 $|$ 57.6 & 40.8 $|$ 33.9 & 10.6 $|$ 8.7 \\
tiger & 88.8 $|$ 69.9 & 68.2 $|$ 51.4 & 25.0 $|$ 19.5 \\
timber wolf & 88.4 $|$ 72.6 & 62.1 $|$ 48.8 & 20.6 $|$ 16.2 \\
tractor & 86.7 $|$ 61.3 & 64.7 $|$ 43.4 & 23.0 $|$ 16.3 \\
train & 75.8 $|$ 51.2 & 52.6 $|$ 35.2 & 21.0 $|$ 13.5 \\
tricycle & 71.2 $|$ 54.2 & 48.7 $|$ 36.2 & 17.8 $|$ 13.2 \\
trolleybus & 82.6 $|$ 60.6 & 71.3 $|$ 46.2 & 37.3 $|$ 22.0 \\
unicycle & 57.6 $|$ 58.1 & 29.1 $|$ 29.6 & 7.4 $|$ 8.0 \\
violin & 75.2 $|$ 50.3 & 52.2 $|$ 26.7 & 19.0 $|$ 9.1 \\
vizsla & 89.1 $|$ 72.4 & 73.9 $|$ 58.3 & 30.0 $|$ 23.5 \\
walker hound & 89.1 $|$ 71.8 & 69.8 $|$ 52.7 & 27.1 $|$ 20.9 \\
warthog & 77.6 $|$ 61.3 & 45.2 $|$ 34.1 & 12.8 $|$ 9.5 \\
washer & 75.0 $|$ 60.6 & 61.0 $|$ 48.0 & 25.9 $|$ 19.5 \\
water buffalo & 73.2 $|$ 54.7 & 37.8 $|$ 27.0 & 8.8 $|$ 6.3 \\
weimaraner & 88.6 $|$ 69.7 & 65.6 $|$ 49.8 & 23.0 $|$ 18.1 \\
wheelchair & 78.4 $|$ 43.1 & 61.3 $|$ 33.6 & 24.6 $|$ 13.5 \\
zebra & 91.3 $|$ 75.1 & 66.0 $|$ 47.2 & 17.9 $|$ 13.1 \\
\bottomrule
\end{tabular}
\end{minipage}
\label{suppl:tab:soc_performance_per_cat}
\end{table*}

\vspace{-15pt}
\section{Example Annotations}
\begin{figure*}[ht]
    \centering
    \includegraphics[width=0.9\linewidth]{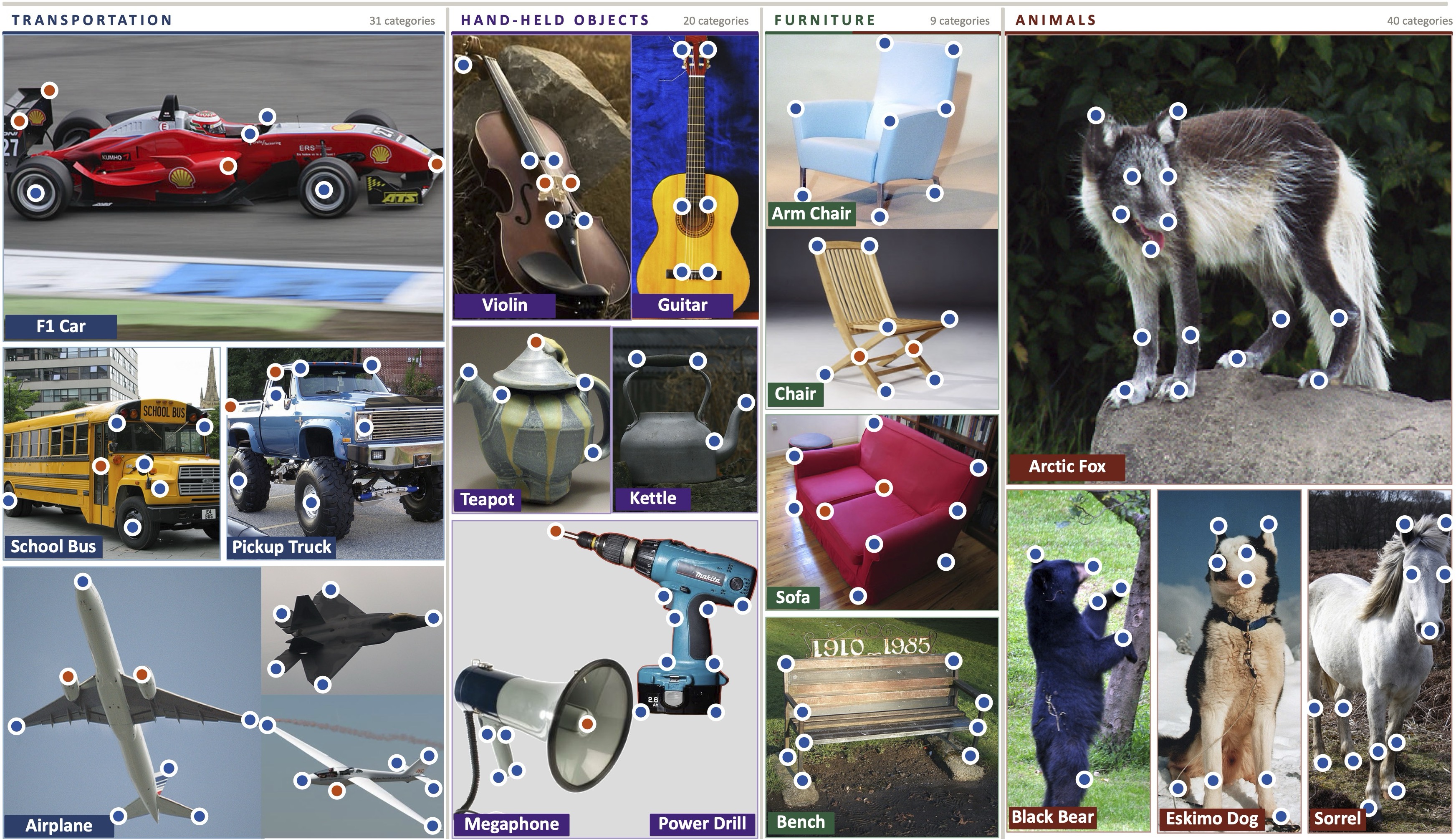}
    
    \caption{\textbf{Example SOCO annotations.}
We visualize example SOCO annotations where a red corresponds to a unique keypoint and blue to a shared keypoint.
    }
    \label{fig:soco_examples}
    
\end{figure*}

We show example annotations in \cref{fig:soco_examples}, illustrating the diversity of the selected categories.
Further, it illustrates keypoints that are unique (red color) and keypoints that are shared across categories or correspond to the same semantic concept.

\subsection{Limitations of Existing SC Annotations}\label{sec:sc_annotation_limitations}

\begin{figure}[t]
    \centering
    \includegraphics[width=1.\linewidth]{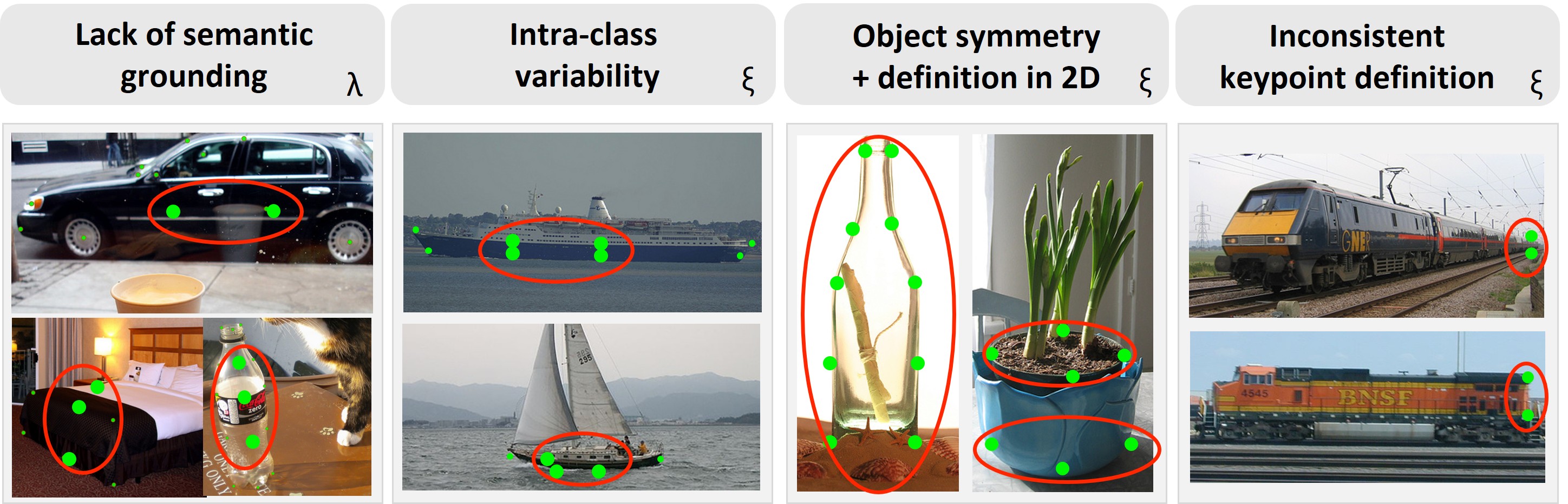}
    \vspace{-10pt}
    \caption{\textbf{Limitations of SC keypoint annotations.}
Current SC datasets include keypoints that lack semantic grounding and are mainly defined geometrically.
This results in particularly ambiguous keypoint definitions for categories with large intra-class variability, e.g., boats.
Uniqueness is not satisfied for symmetric objects where the keypoints are defined via a 2D projection (e.g., for \textit{potted plant} and \textit{bottle}).
Furthermore, some keypoint definitions are inconsistent, for example for \textit{trains}.
Example images are sourced from MISC210K ($\lambda$) and SPair-71k ($\xi$).
    }
    \label{fig:spair_problems}
    \vspace{-10pt}
\end{figure}

\Cref{fig:spair_problems} shows concrete failure cases of keypoint annotations in existing SC benchmarks, illustrating the limitations summarized in \cref{sec:corr:limitations} of the main paper: lack of semantic grounding, intra-class ambiguity, symmetry-induced non-uniqueness, and inconsistent definitions across instances.

\clearpage

\section{Evaluation Results for Other Tasks}

\subsection{Previous SC datasets}
We report evaluation results for other SC datasets in \cref{tab:misc_spair_blink}.
While the rankings for the best models and the worse models remain largely consistent and DINOv2 remains the best-performing model across all datasets, some model rankings change.
For example, I-JEPA is ranked better for SOCO than for MISC or SPair.
One potential explanation for this could be that I-JEPA is trained on ImageNet and SOCO images are also sourced from ImageNet.
On the other hand, PE-Spatial's relative performance drops on SOCO compared to, e.g., SPair.
It is relevant to note that rankings for AP-10K and the SOCO animal subset are largely consistent, as both capture animal keypoint datasets. 
However, rankings change for the whole SOCO dataset, as man-made objects are added.

\begin{table*}[ht]
\caption{
\textbf{Evaluation on other SC benchmarks.}
We report model PCK@0.1 performance with our standard evaluation protocol for MISC210K, SPair-71k, and, AP-10K.
MISC* indicates that we only evaluate the single-instance correspondences, as this is the comparable setting.
While DINOv2 remains the best model, other model rankings vary.
}
\label{tab:misc_spair_blink}
\centering
\small
\setlength{\tabcolsep}{5pt}
\begin{tabular}{lccc}
\toprule
Model & MISC* & SPair & AP-10K \\
\midrule
DINOv1 & 40.05 & 27.01 & 31.87 \\
DINOv2 & 74.86& 58.42& 61.30\\
DINOv3 & 69.71 & 53.44 & 58.57 \\
iBOT & 50.63 & 34.95 & 44.14 \\
I-JEPA & 47.77 & 41.01 & 47.71 \\
C-RADIOv3 & 70.56 & 49.05 & 51.45 \\
DUNE & 57.51 & 42.34 & 47.11 \\
SD 2.1 & 60.15 & 50.71 & 47.55 \\
CroCov2 & 16.01 & 8.41 & 10.00 \\
MAE & 13.46 & 6.62 & 10.08 \\
PIXIO & 52.39 & 37.50 & 39.28 \\
CLIP & 28.30 & 15.44 & 19.58 \\
PE-Spatial & 64.52 & 46.12 & 42.31 \\
QWEN-L & 34.69 & 16.40 & 16.74 \\
\bottomrule
\end{tabular}
\end{table*}

\subsection{Other Downstream Tasks}
\begin{figure}[t]
    \centering
    \includegraphics[width=\linewidth]{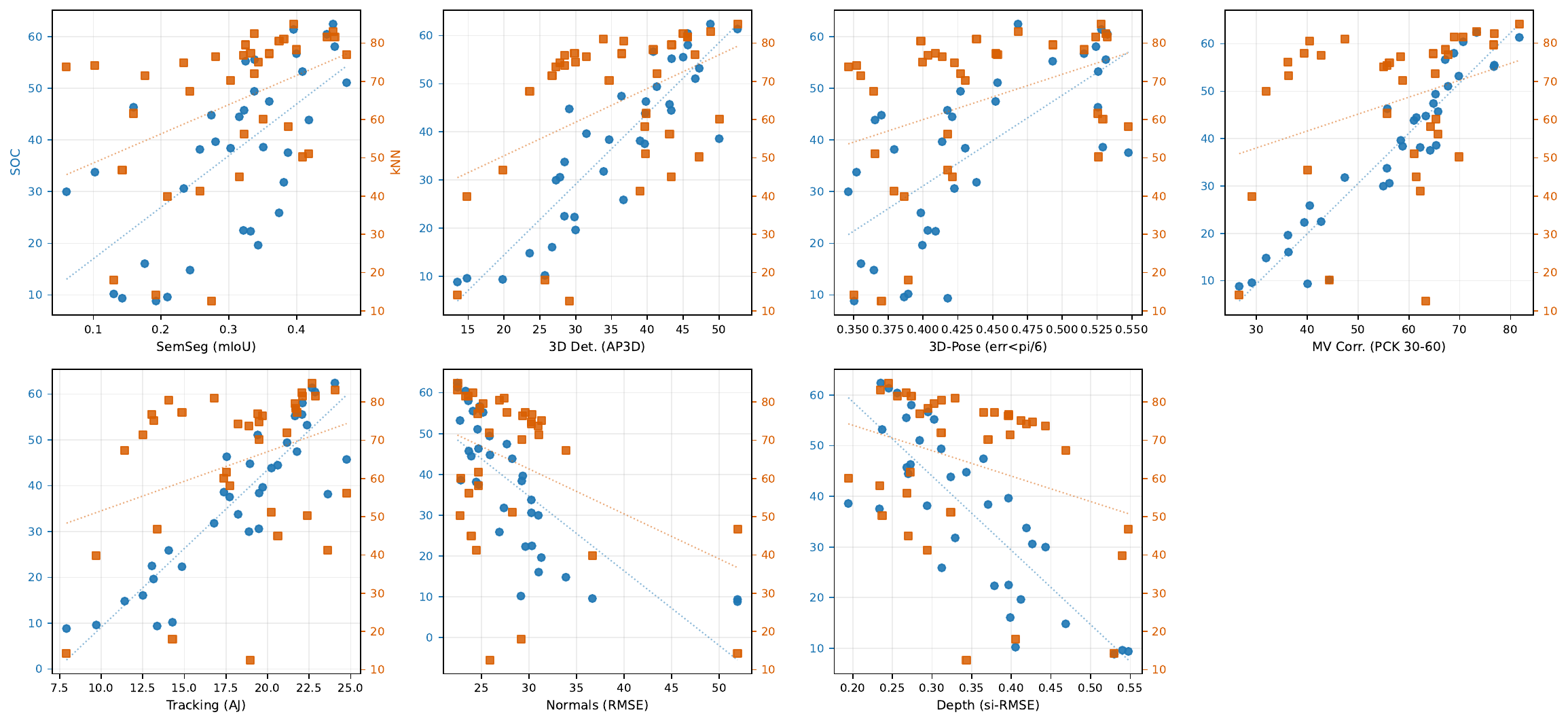}
    \caption{\textbf{SOC and kNN performance vs. downstream task performance.} 
    We plot the model performances for the compared models (as reported in \cref{tab:downstream_tasks}).
    }
    \label{fig:benchmark_vs_soco}
\end{figure}

We report the correlation coefficients and confidence intervals of ImageNet kNN classification / SOC  and other tasks in \cref{tab:downstream_task_correlations_soco_knn}.
Further, we report all results of selected models and datasets in \Cref{tab:downstream_tasks} and visualize them in \cref{fig:benchmark_vs_soco}.

\begin{table*}[t]
\begin{tiny}
\centering
\caption{\textbf{Performance on various downstream tasks.} 
We present the results for various tasks and models, as presented in the main paper.
}
\renewcommand{\arraystretch}{1.05}
\setlength{\tabcolsep}{3pt}

\begin{tabular}{lcccccccc}
\toprule
\multirow{3}{*}{model} &
\textbf{ImageNet} &
\textbf{SemSeg} &
\textbf{3D Det.} &
\textbf{3D-Pose} &
\textbf{MV Corr.} &
\textbf{Tracking} &
\textbf{Normals} &
\textbf{Depth} \\
& kNN & ADE20K & ARKit & ImageNet3D & NAVI & TAP-Vid-D. & NYUv2 & NYUv2 \\
& top-1 $\uparrow$ & mIoU $\uparrow$ & AP3D $\uparrow$ & $\text{err} < \pi/6$ $\uparrow$ & PCK $\theta_{30}^{60}$ $\uparrow$ & AJ $\uparrow$ & RMSE $\downarrow$ & RMSE $\downarrow$ \\
\midrule
DINOv1 & 75.04 & 0.24 & 28.97 & 0.42 & 56.60 & 19.48 & 31.67 & 0.42 \\
DINOv2 & 81.64 & 0.45 & 45.37 & 0.53 & 68.94 & 22.86 & \textbf{23.38} & 0.25 \\
DINOv3 & \textbf{82.37} & 0.42 & \textbf{46.70} & 0.53 & \textbf{73.94} & 22.07 & 24.14 & 0.28 \\
iBOT & 76.30 & 0.29 & 32.96 & 0.41 & 58.09 & 19.71 & 29.23 & 0.39 \\
I-JEPA & 60.51 & 0.18 & 41.45 & 0.53 & 55.44 & 17.55 & 25.13 & 0.28 \\
C-RADIOv3 & 76.95 & \textbf{0.48} & \textbf{46.70} & 0.47 & 66.18 & 19.36 & 24.75 & 0.29 \\
DUNE & 56.61 & 0.33 & 43.10 & 0.42 & 64.03 & \textbf{24.76} & 23.76 & 0.27 \\
SD 2.1 & 4.86 & 0.25 & 30.27 & 0.36 & 58.41 & 15.31 & 27.60 & 0.35 \\
CroCov2 & 18.14 & 0.16 & 26.53 & 0.39 & 45.43 & 14.29 & 29.29 & 0.41 \\
MAE & 46.28 & 0.16 & 20.60 & 0.42 & 40.10 & 13.36 & 51.92 & 0.55 \\
PIXIO & 58.36 & 0.39 & 39.63 & \textbf{0.55} & 60.02 & 17.72 & 25.19 & \textbf{0.24} \\
CLIP & 72.08 & 0.24 & 27.93 & 0.36 & 35.20 & 12.64 & 30.29 & 0.39 \\
PE-Spatial & 53.07 & 0.43 & 40.63 & 0.37 & 60.61 & 20.45 & 26.68 & 0.31 \\
\bottomrule
\end{tabular}

\label{tab:downstream_tasks}
\end{tiny}
\end{table*}

\begin{table}[t]
\caption{\textbf{Correlation coefficients for downstream tasks.}
Pearson correlation coefficients with 95\% confidence intervals between downstream-task performance and SOCO / ImageNet kNN.
The results correspond to the bar plot in the main paper.
}
\label{tab:downstream_task_correlations_soco_knn}
\centering
\footnotesize
\setlength{\tabcolsep}{7pt}
\begin{tabular}{lcccc}
\toprule
 & \multicolumn{2}{c}{SOCO} & \multicolumn{2}{c}{kNN} \\
\cmidrule(lr){2-3} \cmidrule(lr){4-5}
Task & $r$ & 95\% CI & $r$ & 95\% CI \\
\midrule
Seg.      & 0.629  & [0.422, 0.795]   & 0.399  & [0.121, 0.646] \\
Det3      & 0.892  & [0.801, 0.949]   & 0.393  & [0.027, 0.631] \\
Pose      & 0.692  & [0.528, 0.823]   & 0.348  & [0.071, 0.566] \\
MV Corr.  & 0.943  & [0.912, 0.969]   & 0.266  & [-0.087, 0.562] \\
Tracking  & 0.907  & [0.844, 0.956]   & 0.286  & [-0.106, 0.594] \\
Normals   & -0.737 & [-0.867, -0.673] & -0.395 & [-0.698, -0.084] \\
Depth     & -0.798 & [-0.888, -0.668] & -0.298 & [-0.570, -0.069] \\
\bottomrule
\end{tabular}
\end{table}

\section{More Details on the Performed Evaluations}

\begin{table}[t]
\centering
\caption{\textbf{Evaluated visual models.}
We list architecture, supervision type, and pre-training data for the evaluated backbones presented in the main paper.
Whenever possible, we use publicly released checkpoints of comparable scale.}
\label{suppl:tab:backbones}
\renewcommand{\arraystretch}{1.05}
\setlength{\tabcolsep}{4pt}
\begin{scriptsize}
\begin{tabular}{lccc}
\toprule
Model & Architecture & Supervision & Pre-train data \\
\midrule
DINOv1~\cite{caron2021emerging} & ViT-B/16 & SSL & ImageNet-1K \\
DINOv2~\cite{oquab2023dinov2} & ViT-B/14 & SSL & LVD-142M \\
DINOv3~\cite{simeoni2025dinov3} & ViT-B/16 & SSL & LVD-1689M \\
iBOT~\cite{zhou2021ibot} & ViT-B/16 & SSL & ImageNet-1K \\
I-JEPA~\cite{assran2023self} & ViT-H/16 & SSL (JEPA) & ImageNet-1K \\
C-RADIOv3~\cite{heinrich2025radiov25improvedbaselinesagglomerative} & ViT-B/16 & Distill. &NV-CC-T2I-Dataset 700M \\
DUNE~\cite{sariyildiz2025dune} & ViT-B/14 & Distill. & DUNE-20.7M \\
SD 2.1~\cite{rombach2022high} & U-Net & T2I gen. & LAION-5B \\
CroCov2~\cite{croco_v2} & ViT-B/16 & MV SSL & HM3D, ScanNet, etc. \\
MAE~\cite{he2022masked} & ViT-B/16 & SSL (MIM) & ImageNet-1K \\
PIXIO~\cite{yang2025pursuit} & ViT-B/16 & SSL (MIM) & Curated MetaCLIP-2B \\
CLIP~\cite{radford2021learning} & ViT-B/16 & Contrastive & $\sim$400M image--text pairs \\
PE-Spatial~\cite{bolya2025perception} & ViT-G/14 & Contrastive & Curated 5.4B MetaCLIP pairs \\
Qwen2.5-VL~\cite{Qwen2.5-VL} & $~$ViT-H & VLM & $~$4T token multimodal data \\
\bottomrule
\end{tabular}
\end{scriptsize}
\end{table}

\subsection{Details on SOC Evaluation}
We follow the evaluation protocol in Probe3D~\cite{el2024probing} for all semantic correspondence evaluations.
Specifically, we compute PCK@0.1 with bounding box normalization and the per-category PCK using the \textit{per-keypoint} convention, as also applied in other recent works, e.g., \cite{zhang2024telling}.
The final result is computed using the average over categories and we keep a fixed image resolution of 800 pixels.

\subsection{Details on Evaluated Models}
We evaluate a diverse set of visual backbones spanning self-supervised, vision–language, generative, and 3D-aware training regimes, as summarized in \Cref{suppl:tab:backbones}. 
All backbones are kept frozen throughout our experiments. 
Zero-shot settings operate directly on the backbone features without any learnable components, while probed settings attach lightweight task-specific heads (e.g., linear or DPT-style decoders) trained on top of the fixed representations. 
This design ensures that downstream performance reflects differences in representation quality rather than task-specific fine-tuning capacity.
We consistently use timestep 250 for SD 2.1.

\subsection{Details on Other Downstream Tasks}

We extend Probe3D~\cite{el2024probing}, a 3D-awareness evaluation framework into a broader, unified evaluation suite spanning monocular geometry, multi-view correspondence, semantic segmentation, tracking, classification, and 3D detection. 
This section details the tasks, datasets, probe architectures, and evaluation protocols used, as well as the backbone families we evaluate.

\begin{figure}[t]
    \centering
    \includegraphics[width=0.8\linewidth]{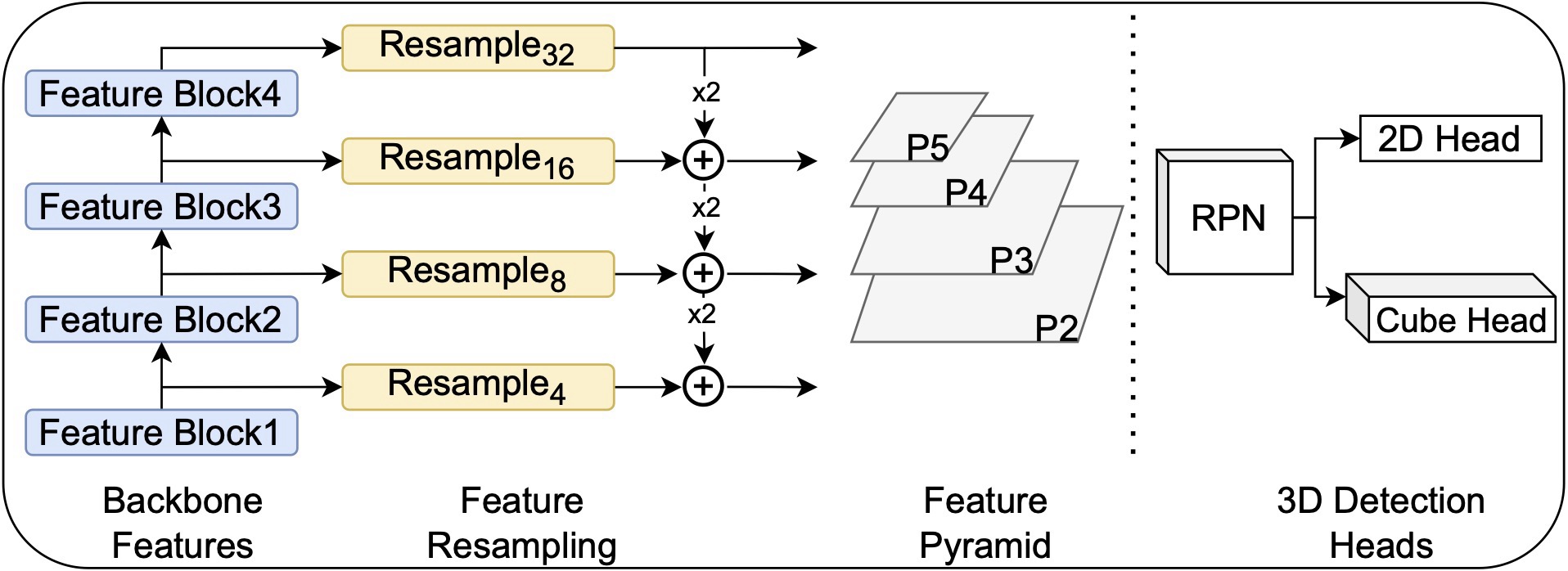}
    \caption{\textbf{Overview of the detection probe.} The probe receives four intermediate feature maps from a pretrained frozen backbone, merges and upsamples them to produce a Detectron2-style feature pyramid \(\{p_2,p_3,p_4,p_5\}\) used by the 3D detection head.}
    \label{fig:dpt_fpn2_arch}
\end{figure}

The extended suite covers the following task families:
\paragraph{Correspondence (zero-shot).}
We evaluate correspondence in two regimes: \textbf{semantic matching} and \textbf{multiview geometric matching}. For SPair-71k~\cite{min2019spair}, we follow Probe3D~\cite{el2024probing} and extract feature vectors at annotated keypoints, predicting matches via nearest-neighbor similarity and reporting PCK@0.1. For NAVI~\cite{jampani2023navi}, we extract feature maps for both views and establish correspondences using nearest-neighbor matching in feature space, followed by Lowe's ratio test to retain reliable matches. Candidate correspondences are triangulated using ground-truth camera calibration, and accuracy is measured as the fraction of matches whose 3D error is below 2\,cm. Following the Probe3D protocol, we stratify the 2\,cm recall by relative camera rotation and report performance in the hardest bin, corresponding to pairs with viewpoint change in the $[90^\circ,120^\circ)$ range.
\paragraph{ImageNet classification (kNN).}
We perform ImageNet classification using $k$-nearest neighbors.
For this embeddings are extracted on the ImageNet training set and evaluated on the validation set.
We select the $k$-value that results in the best classification accuracy.
Following common practice, classification is performed using the CLS token if available. Otherwise, dense tokens are averaged into one vector.
\paragraph{Semantic segmentation (probed).}
Dense semantic understanding is assessed on ADE20K~\cite{zhou2017scene} using a minimal segmentation probe. We train a lightweight linear segmentation head consisting of a single $1\times1$ convolution applied to dense frozen backbone features on ADE20K. The probe is trained for 25 epochs using SGD, and we report mean IoU on the validation set.
\paragraph{Tracking (zero-shot).}
We assess spatio-temporal consistency via zero-shot point tracking on TAP-Vid-DAVIS~\cite{doersch2022tap}. Dense feature maps are extracted for each frame, and query points are embedded by bilinear sampling in feature space at their first visible location. For each subsequent frame, we compute the cosine similarities between the query descriptor and the dense feature map, and obtain correspondences via argmax operation. Evaluation follows the TAP-Vid queried-first protocol, and we report Average Jaccard (AJ)~\cite{aydemir2024visualfoundationmodelsachieve}, which jointly captures occlusion consistency and point localization accuracy.
\paragraph{Monocular geometry (probed).}
We evaluate single-image geometric prediction on NYUv2~\cite{nyudepthECCV12} using two tasks: depth estimation and surface normal prediction. Following the Probe3D~\cite{el2024probing} setup, we attach a lightweight DPT-style multiscale decoder to frozen backbone features extracted from several intermediate blocks. For depth estimation, we use metric depth on NYUv2 and evaluate performance using the root mean squared error (RMSE) between predicted and ground-truth depth maps. For surface normals, the decoder predicts per-pixel normal directions, and accuracy is assessed using the RMSE of angular errors between predicted and ground-truth normals, providing a direct measure of local geometric fidelity.
\paragraph{3D pose estimation (probed).}
We evaluate object-level 3D awareness on ImageNet3D~\cite{ma2024imagenet3d} by linearly probing frozen backbone features for 3D viewpoint prediction. Following the ImageNet3D protocol, three independent linear probes are trained to predict azimuth, elevation, and in-plane rotation from pooled backbone features. The predicted angle distributions are converted to continuous rotation matrices, and performance is measured using the geodesic rotation error~\cite{ma2024imagenet3d}, defined as the angle of the matrix logarithm of $R_{\mathrm{pred}}^\top R_{\mathrm{gt}}$. We report pose accuracy as the percentage of samples whose rotation error is below a threshold of $\pi/6$.
\paragraph{3D detection (probed).}
Our experiments build on the Omni3D~\cite{brazil2023omni3d} detection pipeline, which extends Detectron2~\cite{wu2019detectron2} with Cube R-CNN style 3D cuboid prediction. While the original setup optimizes a CNN backbone, we repurpose it as a 3D detection head on top of frozen, pretrained visual encoders (eg., DINO/v2/v3, CLIP, SD etc.). To bridge the gap between pre-trained backbones and 3D detection heads, we introduce a lightweight DPT~\cite{ranftl2021vision} probe, and the resulting features are reassembled to form a feature pyramid (see~\Cref{fig:dpt_fpn2_arch}). 
Given a pretrained backbone, we select four feature blocks at increasing depth and reshape their patch tokens into dense spatial feature maps. These maps all share the same spatial resolution (the patch grid) but capture progressively higher-level semantics. The four features are fed into the probe, which first unifies channel dimensions with 1×1 convolutions, then constructs a top-down FPN~\cite{lin2017featurepyramidnetworksobject} style decoder.
Through resampling and lateral fusion, the probe produces a Detectron2-compatible feature pyramid. 

We attach the probe and detection heads to frozen backbone features and train on a subset of indoor RGB-D scenes with 3D bounding box annotations. We report average precision (AP3D) for ARKitScenes~\cite{baruch2022arkitscenesdiverserealworlddataset} subset of Omni3D in~\Cref{tab:downstream_tasks}.

\begin{figure}[t]
    \centering
    \includegraphics[width=\linewidth]{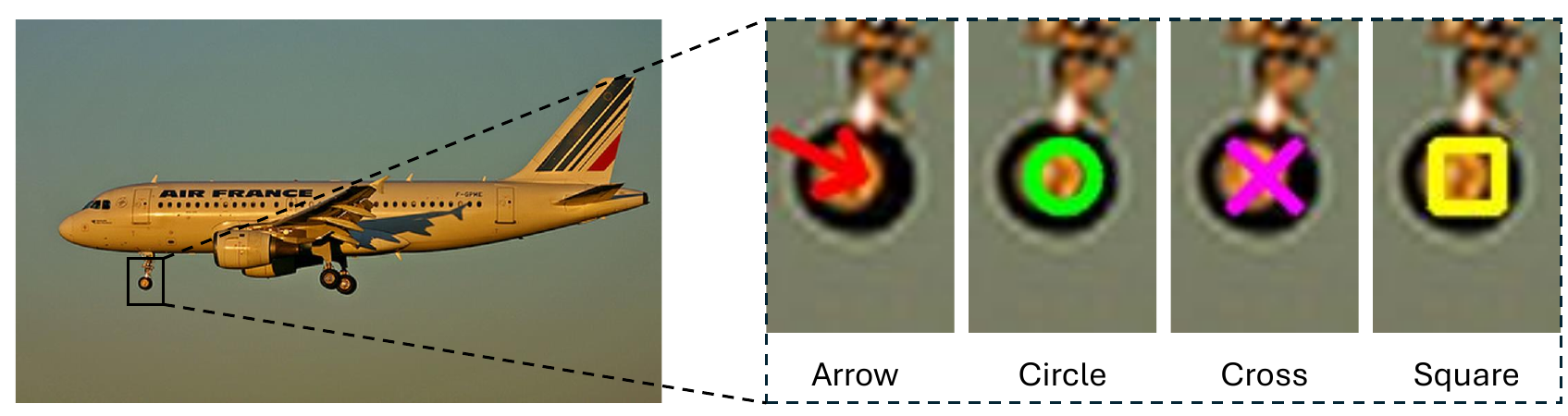}
    \caption{\textbf{Examples of visual markers} to indicate keypoints for LVLM evaluation.}
    \label{fig:lvlm_visual_prompt}
\end{figure}
\begin{table}[t]
    \centering
    \caption{\textbf{Evaluation of visual prompts.}Average performance of Qwen2.5-VL-7B-Instruct given different visual prompts.}
    \begin{minipage}{0.45\linewidth}
        \centering
        \begin{tabular}{lc}
\toprule
Marker shape & Mean\\
\hline
Arrow          &  \textbf{30.0}\\
Circle          & 29.8\\
Cross          & 28.0\\
Square          & 26.6\\

\bottomrule

\end{tabular}
        
        \label{tab:first}
    \end{minipage}%
    \hfill
    \begin{minipage}{0.45\linewidth}
        \centering
        
\begin{tabular}{lc}
        \toprule
        Marker color & Mean\\
        \hline
        Red          & \textbf{30.0}\\
        Blue         & 29.3\\
        Yellow       & 28.9\\
        Purple       & 28.2\\
        Green        & 27.4\\
        \bottomrule
        \end{tabular}
        
    \end{minipage}
    \label{tab:lvlm_visual_prompts_comp}
\end{table}

\begin{figure*}[ht]
    \centering
    \begin{subfigure}[t]{0.3\linewidth}
        \centering
        \includegraphics[width=\linewidth]{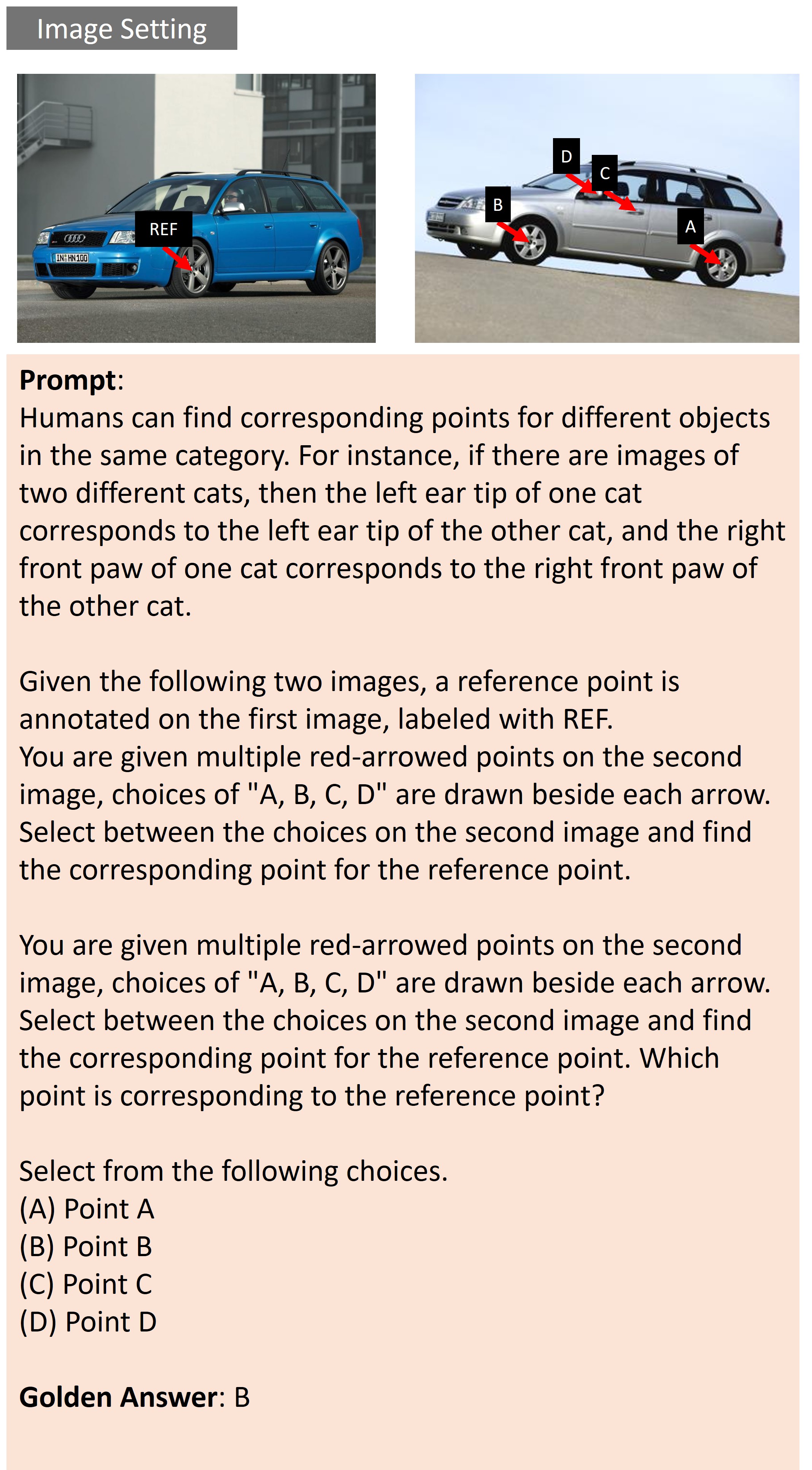}
        \caption{Vis.\ setting.}
        \label{fig:lvlm_prompt_img}
    \end{subfigure}
    \hfill
    \begin{subfigure}[t]{0.3\linewidth}
        \centering
        \includegraphics[width=\linewidth]{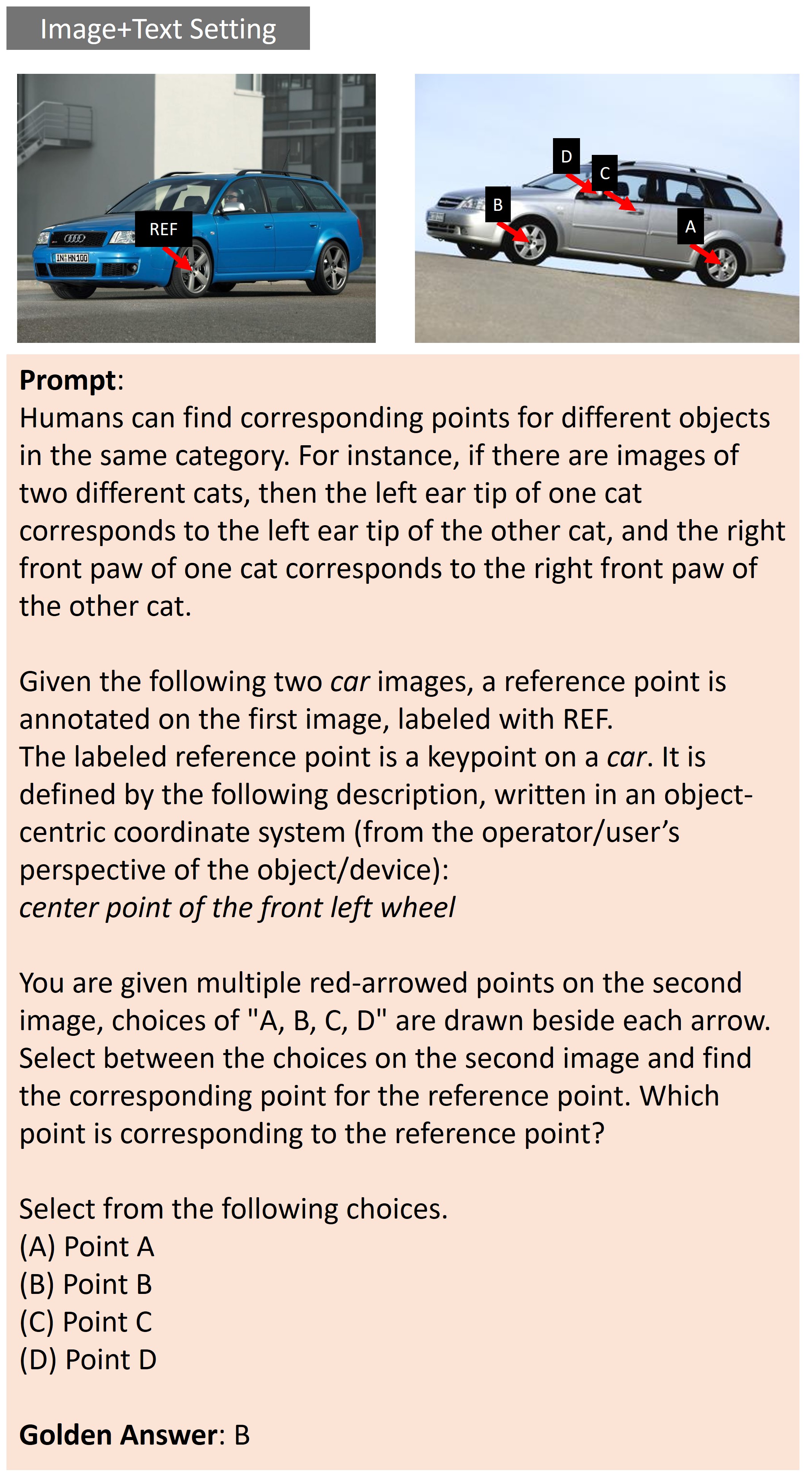}
        \caption{Vis.+Desc.\ setting.}
        \label{fig:lvlm_prompt_imgtxt}
    \end{subfigure}
    \hfill
    \begin{subfigure}[t]{0.3\linewidth}
        \centering
        \includegraphics[width=\linewidth]{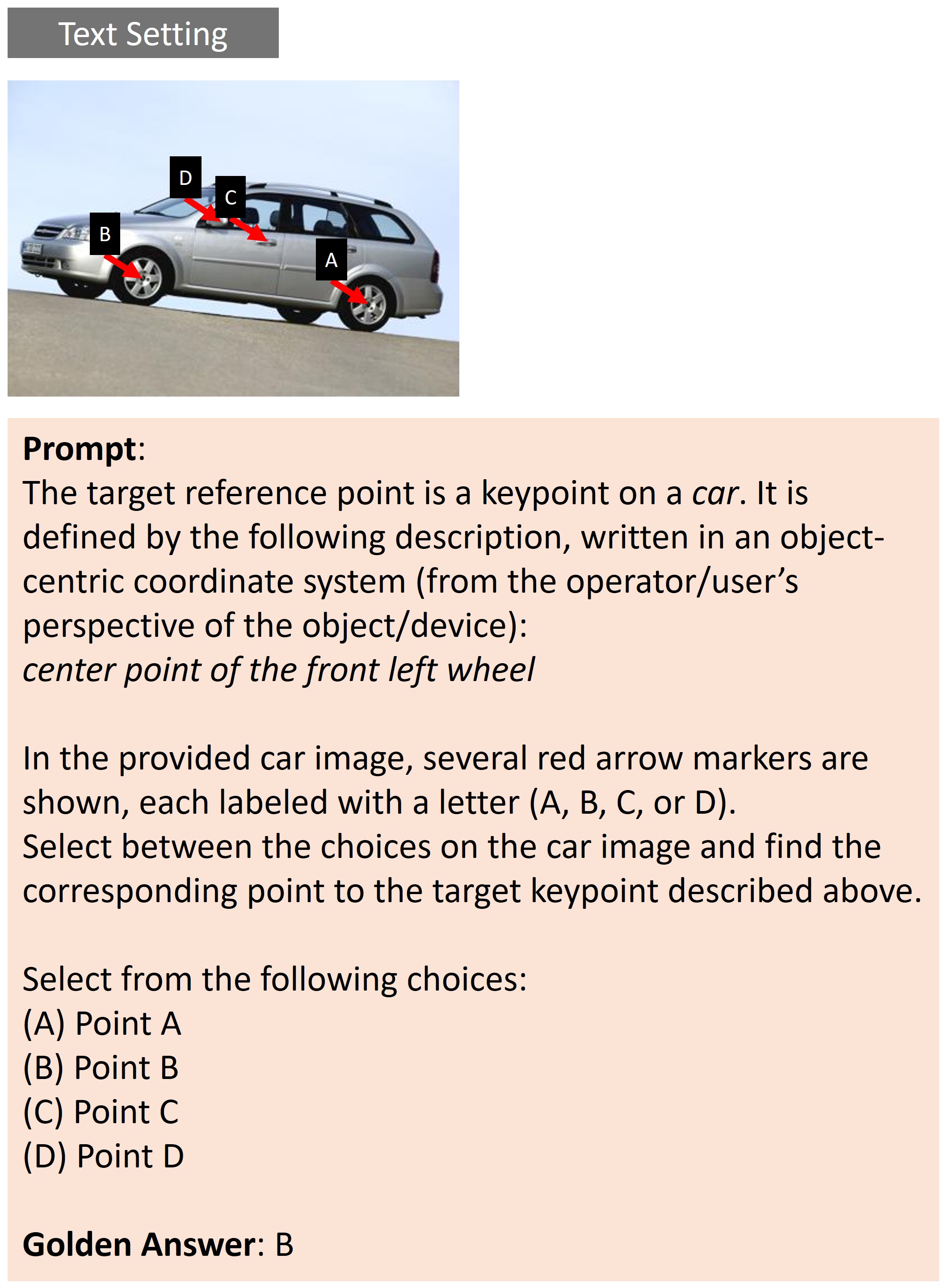}
        \caption{Desc.\ setting.}
        \label{fig:lvlm_prompt_txt}
    \end{subfigure}
    \caption{\textbf{Example prompts for LVLM evaluation} in the three settings \textit{image}, \textit{image+text}, and \textit{text}.}
    \label{fig:lvlm_prompts}
\end{figure*}

\subsection{More Details of LVLM Evaluation}

\textbf{Implementation details.}
We use the VLMEvalKit~\cite{liu2024mmbench} framework to perform standardized evaluation across different LVLMs.
For the evaluation, we pursue a similar setup as \cite{fu2024blink}.
GPT-4o is employed as a judge to verify whether an LVLM's output matches the ground-truth answer.
From the annotated SOCO data, we construct 2{,}000 multiple-choice questions.
For each question, we provide the human-annotated semantically matched keypoint as the ground-truth answer and use other randomly sampled annotated keypoints in the target image as distractor options.

\noindent\textbf{Prompts}: 
We illustrate the evaluation setting and present prompt examples for the LVLM evaluation under different settings in \cref{fig:lvlm_prompt_img}, \cref{fig:lvlm_prompt_txt}, and~\cref{fig:lvlm_prompt_imgtxt}.
For the \emph{Vis.} setting, we provide BLINK-style questions with the red arrow marker in the source image.
For \emph{Vis.+Desc.}, we additionally include a templated keypoint description alongside the visual marker.
For the \emph{Desc.} setting, the source image is omitted and the query keypoint is specified only through its textual description; the target image with candidate markers remains visible.

\noindent\textbf{Choices of visual markers}: 
The BLINK benchmark uses a red circle to mark keypoints for LVLMs to attend to.
Previous work~\cite{shtedritski2023does, cai2025depthlm} has shown that different visual markers can affect VLM performance.
Here, we investigate alternative visual markers for keypoints to study their impact.
We experiment with different colors and shapes of visual markers, with examples shown in Figure~\ref{fig:lvlm_visual_prompt}.

\begin{figure*}[ht]
    \centering
    \includegraphics[height=0.39\textheight, angle=90]{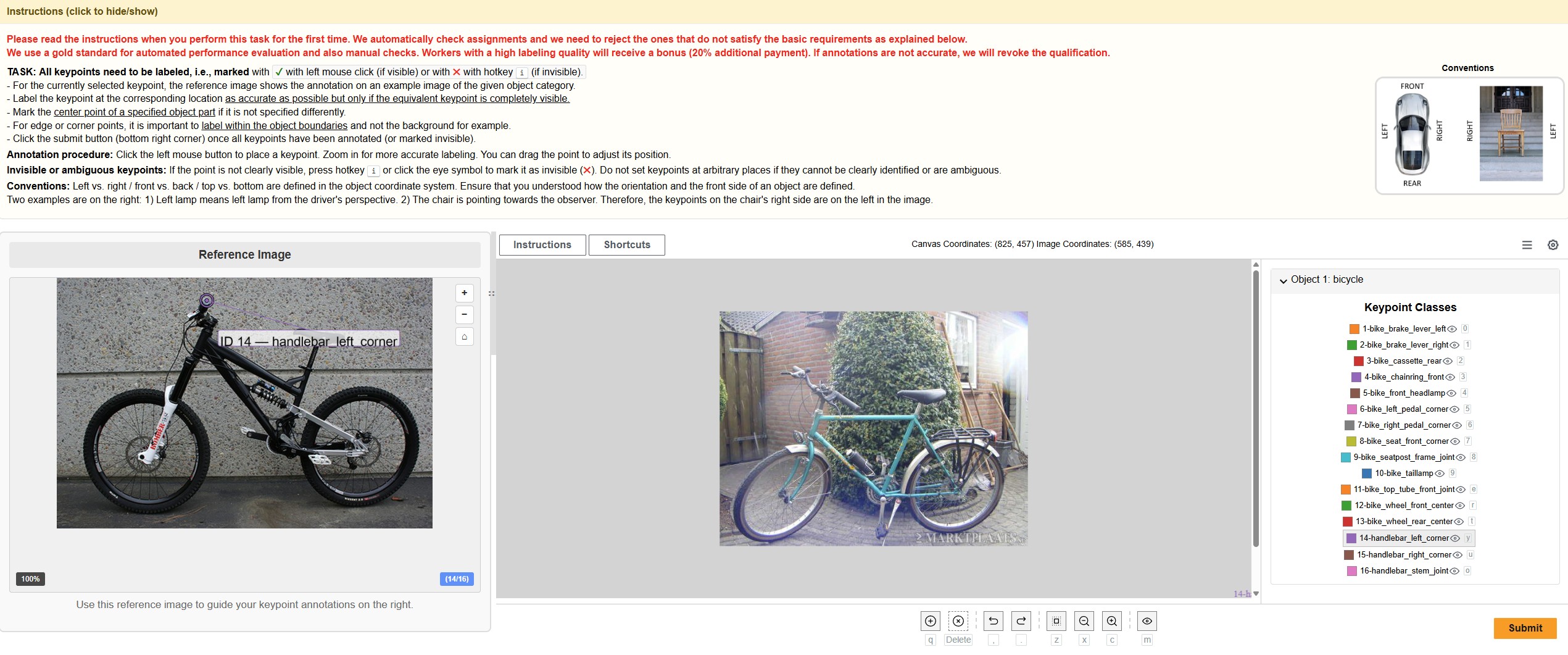}
    \caption{\textbf{Example AMT labeling GUI. }
    This GUI was presented to Amazon Mechanical Turk workers for keypoint labeling.
    }
    \label{fig:gui_amt}
\end{figure*}

To assess how robust LVLMs are to different visual markers, we follow the BLINK benchmark and build a smaller benchmark of SPair-71k to search for markers that yield the highest accuracy. 
Following the BLINK protocol, we construct 233 questions and we present results using Qwen2.5-VL-7B-Instruct across all settings.
\Cref{tab:lvlm_visual_prompts_comp} reports the average performance of the LVLM under different marker shapes and colors.
We observe that the arrow shape and red color achieve the best performance among the tested options respectively.
Consequently, we assume this setting also generalizes to the SOCO dataset and we adopt red arrow as the default visual marker in all LVLM experiments.

\section{More Details about Annotation Pipeline}
\Cref{fig:gui_amt} presents an example GUI that was shown to the AMT workers that were hired for labeling the keypoints.
Reference annotations were given on representative images for each keypoint.
Every AMT worker had to pass a qualification test before annotating and continuous monitoring ensured sufficient labeling quality.

\section{Limitations}

\textbf{Sparse keypoint annotations.}
SOCO provides characteristic part correspondences via sparse keypoint labels rather than dense semantic matching. This is sufficient for diagnosing structured part-level understanding, but it does not support evaluating dense pixel-wise correspondence \textit{per-se}.

\textbf{Image-source bias.}
Images are sourced from ImageNet3D~\cite{ma2024imagenet3d} and Animal3D~\cite{xu2023animal3d}, which enables inherited 3D pose metadata and in-distribution evaluation of ImageNet-trained models but biases the dataset toward salient, curated object views and limits the evaluation of out-of-distribution scenarios.

\textbf{Prompted LVLM setting.}
Keypoint descriptions are template-based.
More detailed natural language descriptions could improve the LVLM performance further.

\textbf{Zero-shot nearest-neighbor matching.}
SOC is mainly designed as a zero-shot diagnostic.
Therefore, the default vision-model evaluation uses nearest-neighbor feature matching, which is intentionally simple and forms a lower bound on what a given representation can support with supervised adaptation. 

\textbf{Cross-category taxonomy scope.}
Cross-category correspondences are defined within the proposed concept hierarchy. Broader functional analogies that fall outside this hierarchy (e.g.\ tool affordance transfer across distant categories) remain future work.

\section{Ethical Concerns}
The SOCO dataset includes a small number of images depicting military equipment (specifically the categories \textit{tank}, \textit{rifle}, and \textit{fighter jet}), but these objects are shown in non-violent contexts and do not directly capture physical harm.
All images were sourced from public datasets~\cite{ma2024imagenet3d,deng2009imagenet}.
The purpose of the dataset is exclusively methodological: to study semantic correspondence and representation learning for diverse categories.
Nonetheless, we acknowledge that models could be evaluated on data containing weapons in principle and could be potentially applied in harmful downstream applications. 

\clearpage